\documentclass{ieeeaccess}
\usepackage{cite}
\usepackage{amsmath,amssymb,amsfonts}
\usepackage{algorithmic}
\usepackage{graphicx}
\usepackage{textcomp}
\usepackage{microtype}
\usepackage{booktabs}
\usepackage{adjustbox}

\usepackage{tabularx}
\usepackage{caption}
\usepackage{url}  
 \usepackage{array,multirow} 
\usepackage{tabularx,ragged2e,booktabs,caption,array,multirow,multicol}

\usepackage{cite} 
 
\usepackage{textcomp,marvosym}

\usepackage{subfigure}
\usepackage{lineno,hyperref}

\usepackage{amssymb}
\usepackage{amsmath}
\usepackage{siunitx}
\usepackage{tabularx}
\usepackage{algorithmic}

\usepackage[utf8x]{inputenc}

\def\BibTeX{{\rm B\kern-.05em{\sc i\kern-.025em b}\kern-.08em
    T\kern-.1667em\lower.7ex\hbox{E}\kern-.125emX}}
    
\begin{document}
\history{Date of publication xxxx 00, 0000, date of current version xxxx 00, 0000.}
\doi{10.1109/ACCESS.2017.DOI}

\title{Semantic and sentiment analysis of selected Bhagavad Gita translations using BERT-based language framework}
\author{\uppercase{Rohitash Chandra *}\authorrefmark{1,2}, \IEEEmembership{Senior Member, IEEE},
\uppercase{Venkatesh Kulkarni *\authorrefmark{3} }}
\address[1]{Transitional Artificial Intelligence Research Group, School of Mathematics and Statistics, University of New South Wales, Sydney,   Australia  }
\address[2]{UNSW Data Science Hub, University of New South Wales, Sydney,   Australia  }
\address[3]{Department of Mechanical Engineering, Indian Institute of Technology Guwahati, Assam, India} 
\tfootnote{* Both authors contributed equally}

\corresp{Corresponding author: R. Chandra (e-mail: rohitash.chandra@unsw.edu.au).}

\begin{abstract} 
It is well known that translations of songs and  poems not only break rhythm and rhyming patterns, but can also result in  loss of semantic information. The Bhagavad Gita is an ancient Hindu philosophical text originally  written in Sanskrit that features a conversation between Lord Krishna and Arjuna prior to the Mahabharata war.   The Bhagavad Gita is also one of the key sacred texts in Hinduism and is known as  the forefront of the Vedic corpus of Hinduism.  In the last two centuries, there has been  a lot of interest in Hindu philosophy from western scholars; hence, the Bhagavad Gita has been translated in a number of languages. However, there is not much work that validates the quality of the English translations. Recent progress of language models powered by deep learning has enabled not only translations but a better understanding of language and texts with semantic and sentiment analysis. Our work is motivated by the recent progress of language models powered by deep learning methods. In this paper, we present a framework that compares selected translations (from Sanskrit to English) of the Bhagavad Gita using semantic and sentiment analyses.  We use hand-labelled  sentiment dataset for tuning state-of-art deep learning-based language  model known as \textit{bidirectional encoder representations from transformers} (BERT).  We  provide sentiment and semantic analysis for selected chapters and verses across translations.  Our results show that although the style and vocabulary in the respective translations vary widely, the sentiment analysis and semantic similarity shows that the message conveyed are mostly similar.

\end{abstract}


\titlepgskip=-15pt

\maketitle

\section{Introduction}

Comparative religion   is a field that encompasses a systematic comparison of the doctrines and practices,   philosophy, theology,  and spiritual viewpoints  of the world's religions \cite{paden2005comparative,sharpe2003comparative,eliade1996patterns}. The area of \textit{philosophy of religion} \cite{meister2009introducing,reese1996dictionary} encompasses the study of central themes and ideas in religions  and the associated cultural traditions that relate to philosophical topics such as metaphysics, epistemology, and ethics. On the other hand, \textit{religious philosophy} \cite{wildman2010religious} is an area that encompasses   philosophical topics but from the view of the religion of interest which is particularly driven by believers, such as Hindu philosophy, Christian theology, Jewish and  Islamic philosophy. However, at times, both groups of scholars can contribute to the subject. Hindu philosophy \cite{bernard1999hindu,saksena1939nature,chaudhuri1954concept} features  central themes of ethics \cite{roy2007just}, nature of consciousness \cite{saksena1939nature}, karma (duty and action) \cite{reichenbach1990law,mulla2006karma},  and the quest for understanding ultimate reality known as Brahman \cite{chaudhuri1954concept}. Hindu philosophy  is also referred to as the Indian philosophy \cite{Dasgupta}\cite{radhakrishnan}. The Upanishads are ancient texts of Hindu philosophy  which have been the foundations of six major theistic schools, where \textit{vedanta} \cite{torwesten1991vedanta} and \textit{yoga} \cite{maas2013concise} have been the most prominent. Hindu philosophy   has  an intersection between philosophy of religion and religious philosophy which has been given a diverse study and examination by western scholars  apart from  Hindu philosophers \cite{historyUpanisahadtrans} \cite{radhakrishnan}. 
 
The Bhagavad Gita is considered as one of the main holy texts of  Hinduism which is part of the Mahabharata which  is   one of the oldest and largest epics written in verse form \cite{rajagopalachari1970mahabharata,hiltebeitel1976ritual,gandhi2010bhagavad}. The Bhagavad Gita, which translates as the song of God  is also known as a text that summarizes and comprises the core of  Hindu philosophy \cite{radhakrishnan}.  In the Mahabharata war, Arjuna is the lead archer   with Lord Krishna as his charioteer representing the Pandava army who are at war with the Kaurava. Arjuna is facing some of  his family members on the other side of the war and becomes confused whether he should go to war or renunciate and become a yogi.  The Bhagavad Gita provides a discussion duty and ethics  as a set of questions from Arjuna and answers by Lord Krishna. The philosophy of karma, known as  karma yoga is attributed to the Bhagavad Gita   \cite{
brown1958philosophy,phillips2009yoga,muniapan2013dharma}. The transmission of the  Bhagavad Gita has been through oral tradition for thousands of years, from the event of the Mahabharata war in India, with recent estimates  around 5100 years ago\cite{namah2019fixing,kak2012mahabharata}; however, the exact dates have been debated in the literature with some scholars dating it to around 2500 - 3500 years ago \cite{gangopadhyayhistoricity,murthy2003questionable}.  We need to note that scholars believe that the Mahabharata was written in ancient Sanskrit where as the Bhagavad Gita has been written in modern Sanskrit; hence, some scholars argue that Bhagavad Gita was later inserted into the epic Mahabharata \cite{minor1986modern,gandhi2010bhagavad}. The original text, which has been composed in Sanskrit, has been ever since translated into various languages. The Bhagavad Gita consists of 18 chapters which feature a series of questions and answers between Lord Krishna and Arjuna with a range of topics that includes the philosophy of Karma. The Mahabharata war lasted for 18 days and hence the organisation of the Gita is symbolic.  The Bhagavad Gita is written in a poetic style so that it can be sung and remembered for generations given the absence of a writing system.  
 Over the last few centuries, there has been a lot of effort in translation of the Bhagavad Gita. There has been a major focus on English translation and commentary by philosophers and scholars of comparative religion, particularly from India, Europe and United States.

Religious linguistics refers to the study of language used in the domain of religion and primarily aims at identifying the structures and functions of religious language (lexicon, syntax, phonology, morphology, etc).  In principle, religious subject matters could encompass a variety of agents, states of affairs or properties, such as God, deities, angels, miracles, redemption, grace, holiness, sinfulness \cite{sep-religious-language}. Recent research suggests that a register of religious language, such as prayer, induces certain psychological effects in the brain  \cite{religious_linguistics} which makes certain components of religious language  a cognitive register. Analysis of religious texts has been prominent, as highlighted, and most translations of the Bhagavad Gita and related texts come with interpretations and commentary regarding philosophy and  how the verses relate to issues at present \cite{theodor2016exploring}. The Bhagavad Gita has been predominantly used as a text and inspired interpretations from areas of  self-help and management, psychology \cite{
jeste2008comparison},  economics \cite{pandey2017economic},   politics \cite{kapila2013political} and leadership\cite{nayak2018effective}. However, these commentaries dwell with the general message of the texts rather than semantic or syntactic analysis. Stein \cite{stein2012multi} extracted 14 multi-word expressions using local grammar and semantic classes from Bhagavad Gita capturing some central themes in the text. Bawa et al. \cite{bawa2021comprehensive} provided a review  on machine translation for English, Hindi and Sanskrit languages. Rajput et al. \cite{rajput2019statistical} performed a statistical analysis of the Bhagavad Gita text for its translations in four languages: English, French, Hindi and Sanskrit. These works, have focused on linguistic and statistical measures which opens the road for machine learning and natural language processing for analysis of religious and philosophical texts.

Natural language processing (NLP) is an area of artificial intelligence that focuses on processing and modelling language  \cite{indurkhya2010handbook,manning1999foundations,chowdhury2003natural}. NLP is typically implemented with language models where deep learning has been prominent for tasks such as topic modelling, language translation, speech recognition, semantic and sentiment analysis \cite{manning1999foundations}. Sentiment analysis  provides an understanding of human emotions  and affective states   \cite{liu2012survey,medhat2014sentiment,hussein2018survey} which has been prominent in understanding customer behaviour \cite{ordenes2014analyzing},  health and medicine  \cite{greaves2013use},  stock market predictions  \cite{mittal2012stock}, and modelling  election outcomes \cite{wang2012system,chandra2021biden}.     Deep learning methods such as long-short term memory (LSTM) network models have been prominently used as language models due to their capability to model temporal sequences with long-term dependencies \cite{hochreiter1997long}. Attention based mechanism in LSTM models have shown significant improvements   \cite{wang2016attention}, and further progress has been made  using \textit{Transformer}  models which  incorporate attention  into  encoder-decoder LSTM framework   \cite{vaswani2017attention,wolf2020transformers}.  The \textit{bidirectional encoder representations from Transformer} (BERT) \cite{devlin2018bert} model  has become a prominent pre-trained language model with knowledge from a large data corpus. BERT features more than 300 million model  parameters for language modeling tasks and  has the potential to provide good semantic and sentiment analysis. 
BERT employs masked language modeling (MLM) for pre-training and has been one of the most successful pre-trained models.
In our earlier work, we used BERT-based framework  for   sentiment analysis of COVID-19 related tweets during the rise of novel cases in India \cite{chandra2021covid}. We also used BERT  for modelling US 2020 presidential elections with sentiment analysis from tweets to predict state-wise winners \cite{chandra2021biden}. Thus, BERT can be used for better understanding the difference using sentiment analysis amongst  translations in philosophical texts such as the Bhagavad Gita. 
 
 The Upanishads are a  set of Hindu philosophical and sacred  texts of which some  are    related  to the Bhagavad Gita in terms of   topics and style. The Upanishads  have been translated more than 150 times from 1783 to 1994, where western scholars make up a significant portion of translators and commentators \cite{historyUpanisahadtrans}. This shows the impact and interest of Hindu texts in the west; hence, analysing the style and quality   of the  translations is essential. It is important to evaluate some of the key translations with the others in terms of sentiments and semantics. The Bhagavad Gita, has been written originally in verse (as a song), and it is well  known that translating such literary works not only breaks  rhyming patterns, but can also change the meaning and message behind the verses \cite{apter2016translating}.

This paper  is motivated by the recent progress of language models powered by deep learning models.  We compare selected translations of the Bhagavad Gita by focusing on sentiment analysis and semantic similarity of the verses. We compare selected translations with a focus on translations directly from Sanskrit to English  and one that uses an intermediate language (Gujarati).  We use novel sentence embedding based on BERT which provides semantic analyses by encoding and representing the verses using sentence embedded vectors.   In the case of sentiment analysis, the pre-trained BERT model is further tuned using a hand labelled Twitter-based training dataset that captures 10 sentiments. We visualise  selected Bhagavad Gita chapter-wise and verse-wise sentiments across the selected translations.

The rest of the paper is organised as follows.  In Section 2, we provide further details about background and historical development behind the composition of the Bhagavad Gita. Section 3 presents the methodology with details regarding data processing and model development.  Section 4 presents the results that feature data analysis and model prediction. Section 5 provides a discussion and Section 6 concludes the paper.

\section{Background}

 \subsection{Historical developments}
 
 The information in Hindu  texts was passed through oral transmission for thousands of years before being written in Sanskrit \cite{graham1993beyond,rocher1993law,parry1985brahmanical}. Although most of the  ancient Hindu texts (Vedas, Upanishads and Bhagavad Gita) used Sanskrit, there are some prominent texts that were written in Tamil such as the Tirukkuṟaḷ \cite{cutler1992interpreting}. Knowledge about natural and astronomical events have been preserved through oral transitions \cite{kak2012mahabharata} which are linked to the Indus valley civilisation (Harappan) with mature period dated 8000 - 9000 years BCE  (before the common era) by recent studies \cite{sarkar2016oxygen}. There are studies that show that the earliest text of Hinduism, known as the Rig Veda, contains  accounts of the Saraswati River \cite{dikshit2012rise} being active. Latest research shows that the Sarawati River started  building sediments and  drying   10,000 years ago \cite{khonde2017tracing}.  The written history and dating  of the composition of the  texts are  continuously being  updated, since the transmission of knowledge was done orally for thousands of years before being written.   There has also been a call to rename the Indus valley civilisation as the Saraswati civilisation and update the dating of the Vedic period \footnote{ 
Chronology of Indian civilisation is ‘dubious’, says IIT-Kharagpur calendar: \url{https://indianexpress.com/article/cities/kolkata/chronology-of-indian-civilisation-is-dubious-says-iit-kharagpur-calendar-7690601/}}. A major challenge has been in translating Vedic (ancient) Sanskrit which is the language of the Rig Veda \cite{petersen1912vedic,hock1975substratum}. The Bhagavad Gita is written in modern Sanskrit which has been easier to translate \cite{johannes2001traditional} when compared to the Rig Veda \cite{aurobindo2018secret}.

\subsection{Bhagavad Gita}


 The Bhagavad Gita features a discourse on duty and karma between Arjuna and Lord Krishna at the battlefield of Kurukshetra. It describes an event at the beginning of the Mahabharata war where Arjuna, the lead archer of the land, has to fight his close family members for the rightful heir to the throne of Hastinapur \cite{priyank2018evaluation} which is now modern day Delhi. Arjuna is the lead archer on the side of the Pandavas with Lord Krishna as his charioteer, who is also a close friend and relative. The Pandavas went on war with the Kauravas who were their close relatives (cousins) for right to the throne of Hastinapur and Indraprastha. Arjuna, after facing his relatives, friends  and gurus on the opposing  side of the battlefield  gets deeply disturbed and emotional, and begins pondering about philosophy of life and essence of dharma (ethics). Arjuna makes a decision not to go to war and  asks a series of questions to his charioteer and mentor, Lord Krishna  who is known as the avatar (born as human) of Lord Vishnu \cite{ShankaraArjuna}, who in the western religious notion would be equivalent to the term God. Due to the complexity of the situation which Arjuna faces about duty and ethics, the set of questions from Arjuna and answers by Lord Krishna makes Bhagavad Gita  the forefront of Hindu philosophy. Bhagavad Gita  also is known as the central sacred text of Hinduism \cite{
bazaz1975role,robinson2014interpretations}.

 \subsection{The Sanskrit language}

 Sanskrit is a  language known for the richness of its vocabulary and layered meanings which induces subtle differences when translated to other languages \cite{bhate1991,burrow2001sanskrit}. Sanskrit is the language for the key Hindu texts such as the Bhagavad Gita, Ramayana, Upanishads, Vedas, and Mahabharata \cite{whitney1884study,van1994sati}. There are a number of  challenges in translating Sanskrit  texts into other languages as discussed next. Firstly,  Sanskrit is  a language for research and scholarship and study of Hinduism; however, it  has a limited number of speakers. Secondly, although the key Hindu mantras for major events  are written in Sanskrit \cite{yelle2004explaining},  majority of Hindus do not understand Sanskrit. On the bright side, Hindi is based on Sanskrit and there is significant vocabulary in Hindi  from Sanskrit \cite{bhatia1987history}. Finally, Sanskrit is a major branch  of Indo-European \cite{gamkrelidze1990early} languages which have links to European languages such as Latin, German and English   \cite{schleicher1874compendium,prichard1857eastern}. Indian Invasion Theory (IIT) proposed  that development of Sanskrit and Hindu (Vedic) texts in ancient India is from invasion/migration of Indo-Aryan tribes into India in pre-Vedic age \cite{thapar1996theory}. IIT has been the forefront of colonialism and has been used as a way for British rulers to motivate their rule and supremacy in British India \cite{cohn1996colonialism}. IIT aligned well with Eurocentric   education system which did not give importance to ancient Indian and Hindu scholars such as Aryabhata who was instrumental in developing the Hindu decimal number system  \cite{gongol2003aryabhatiya,kaye1919indian,ganguli1927elder},  which was branded as Arabic numerals  \cite{vowles1934introduction}.  IIT has been contested heavily   \cite{frawley1994myth,prasanna2012there} with evidence from Harrapan excavations \cite{sarkar2016oxygen}   due to inconsistent timelines  about migration/invasion and lack of evidence  given by IIT. Apart from this, the Rig Veda which is the earliest Hindu text features hymns of Saraswati River \cite{sankaran1999saraswati} when it was flowing  more than 10,000 years ago \cite{khonde2017tracing}. There is an  opposing Out of India Theory (OIT)   \cite{chakrabarty2008public} that suggests that the evolution of Sanskrit and related languages moved from India to Europe and other parts of Asia through migrations.
 
 P{\=a}nini who is known as the 'father of linguistics" \cite{kiparsky1995painian} was a revered philologist and grammarian in ancient India \cite{emeneau1955india}. P{\=a}nini is known for his text  Ast{\=a}dhy{\=a}y{\=\i}\cite{katre1989} 
 which provides a treatise on Sanskrit grammar with rules on linguistics, syntax and semantics covered in verse form. The Sanskrit language has over the decade gained a lot of interest in computational linguistics and computer science. There has been work done in understanding Sanskrit from the view of language models and natural language processing  \cite{briggs1985knowledge,kak1987paninian,kak1987paninian}. Briggs \cite{briggs1985knowledge} argued that Sanskrit is the best language for modelling using artificial intelligence methods. 
 
  There are more than 300 English translations of the Bhagavad Gita in English. An earlier review provided a chronology of translations and commentaries mostly from western scholars from 1785 to 1980 \cite{larson1981song}. The Bhagavad Gita has been translated in more than 100 languages.

 \section{Methodology} 
 
 \subsection{Datasets: Translations of the Bhagavad Gita}

 We select  prominent translations of the Bhagavad Gita that span different translation times-frames (decades) and focus on scholars from a Hindu background in order to avoid any translation biases \cite{gandhi2010bhagavad}. Our framework for analysis of translations can be extended to other translations of the Bhagavad Gita, we merely select these texts as a proof-of-concept. We tried to use the older and the most popular translations, ensuring that the translations  feature verse by verse translations.  The translations  typically contain an interpretation which have been discarded in our analysis. 
 
\begin{table*}[htbp!]
\small
\centering
\begin{tabular}{l l l }
\hline
Texts & Translator & Year\\ \hline
\hline
The Gita according to Gandhi\cite{desai2020gita} & Mahatma Gandhi \& Mahadev Desai &  1946\\ 
The Bhagavad Gita\cite{easwaran2007bhagavad} &Eknath Easwaran &  1985\\ 
The Bhagavad Gita\cite{swami_purohit} & Shri Purohit Swami & 1935\\ 
\hline
\end{tabular}
\caption{Details of the texts used for sentiment and semantic analysis.}
\label{tab:texts}
\end{table*} 

The Bhagavad Gita consist of 18 chapters which features a series of questions and answers between Lord Krishna and Arjuna with a range of topics that includes the philosophy of Karma. The Mahabharata war lasted 18 days \cite{thapar2009war} and hence the organisation of the Gita is symbolic. In our work, we consider three prominent  translations of Bhagavad Gita as the   source of data for analysis. The translators' name, text name,  and the year of  publication  are shown in the Table \ref{tab:texts}. 
 
 The English translation of the Bhagavad Gita published in 1946 by Mahadev Desai \cite{desai2020gita} was translated from Gujarati version based on the commentary known as "Anasakti Yoga" by Mahatma Gandhi \cite{gandhi2010bhagavad}. It is based on talks given by Gandhi between February and November 1926 at the Satyagraha Ashram in Ahmedabad, India. This was the time when Gandhi had withdrawn from mass political activity and devoted much of his time and energy to translating Gita from Sanskrit to his native language Gujarati. As a result, he met with his followers daily, after morning prayer sessions, to discuss the Gita's contents and meaning as it unfolded before him.   Mahatma Gandhi  viewed the Bhagavad Gita not as a historical work, but one that describes the perpetual conflict that rages on in the hearts of the mankind  under the guise of physical warfare. Hence, Gandhi's non-violent interpretation is in stark contrast with contemporary Indian freedom fighters, such as  Lala Lajpat Rai, Aurobindo Ghose, Bal Gangadhar Tilak. These freedom fighters   argued that it was the sacred duty of Indians to fight against the British colonial rule, just as Krishna had urged Arjuna to fulfill his sacred duty \cite{mclain2019living}. The translation by Mahatma Gandhi is accompanied by short commentaries in every chapter which are discarded in our analysis.

The Bhagavad Gita translation by Eknath Easwaran is a direct Sanskrit-English translation published in 1985. Easwaran's comprehensive introduction (55-page)  places the Bhagavad Gita in its historical setting, and brings out the universality and timelessness of its teachings. It features chapter introductions to clarify key concepts and provides a context, and notes and a glossary to explain Sanskrit terms. Eknath Easwaran's commentary  shows the Gita’s relevance to the modern society and highlights key topics in ethics, humanism  and spirituality. Note that the introduction and chapter commentaries have been discarded in our study. The Bhagavad Gita translation by  Shri Purohit Swami \cite{swami_purohit} avoids the use of Sanskrit concepts that may be unfamiliar to English-speakers, for example translating the word 'yoga' as 'spirituality'. However, we note that 'yoga' has been translated as 'union' in prominent works.  This is also a  direct Sanskrit- English translation published much earlier than the previous ones (1935). Unlike Eknath Easwaran, Shri Purohit Swami does not present a commentary on the Bhagavad Gita, but offers direct translation of the text.

We also considered other prominent translations and commentaries, such as those by A. C. Bhaktivedanta Swami Prabhupada \cite{prabhupada}, Paramhansa Yogananda \cite{yogananda}, Sarvepalli Radhakrishnan \cite{radhakrishnan1949bhagavadgita}, and Swami Chinmayananda Saraswati \cite{chinmayananda}; however, due to limited scope of our study, we could not use them. These texts can be considered in future studies.  
 
 \subsection{Text data extraction and processing}
 
 In order to process the respective   files given in printable document format (PDF), we converted them into text files. This conversion from PDF to text file gave us a raw dataset consisting of all the texts shown in Table \ref{tab:preprocessed_texts}. Next, pre-processing is done on the entire dataset, which included the following steps.
\begin{enumerate} 
    \item Removing unicode characters generated in the text files due to noise in the PDF files;
    \item Removing verse numbering in   the Bhagavad Gita;
    \item Replacing the archaic words such as "thy" and "thou" with  modern words for retaining semantic information;
    \item Replacing different words used to refer to  the respective protagonists, Krishna and Arjuna (Table \ref{table:wordconvert});
    \item Converting verses to a single line. 
    \item Removing commentary on the verses by the author.
    \item Removing repetitive and redundant sentences such as "End of the Commentary".
\end{enumerate}

 Table \ref{table:wordconvert} shows the conversion of archaic words and aliases used for Arjuna and Krishna. In Mahatma Gandhi's translation, there are 136 occurrences of different words used to refer to Arjuna, some of them being - "Partha", "Kaunteya", "Bharata", "Mahabahu", etc. The words such as  “Savyasachin”, which literally means ambidextrous, have been replaced according to the context in which they appeared (in this case -- Arjuna, because he could shoot arrows with both hands). Similar conversions have been made for different names of Krishna. In Shri Purohit Swami's translation, Krishna is more commonly referred to as the "lord" (101 occurrences). These word transformations are done to enable the language model to decipher the meaning and context from the verses easily. In the case of the translation by Eknath Easwaran, no archaic words were used and also different names were not used for Krishna and Arjuna; hence, Table \ref{table:wordconvert} does not provide any information.

 \begin{table*}[htbp!]
 \small
\centering  
\begin{tabular}{ c  c  c  c c} 
    \hline\hline 
    Original Word & Transformed word & Mahatma Gandhi & Shri Purohit Swami &  Eknath Easwaran\\ [1ex] 
    \hline 
    "thou" & "you" & 134 & 118 & -\\ [0.5ex] 
    thy & your & 84 & 84 & - \\ [0.5ex]
    thee & you & 60 & 58 & - \\ [0.5ex]
    partha & arjuna & 38 & 0 & - \\ [0.5ex]
    art & are & 26 & 20 & - \\ [0.5ex]
    shalt & shall & 21 & 23 & - \\ [0.5ex]
    kaunteya & arjuna & 21 & 0 & -\\ [0.5ex]
    bharata & arjuna & 19 & 0 & -\\ [0.5ex]
    thyself & yourself & 11 & 11 & -\\ [0.5ex]
    Hrishikesha & Krishna & 7 & 0 & -\\ [1ex] 
    \hline 
\end{tabular}
\caption{Text processing using word conversion with same meaning. }
\label{table:wordconvert} 
\end{table*}

\begin{figure}
    \centering
    \includegraphics[width=7.8cm]{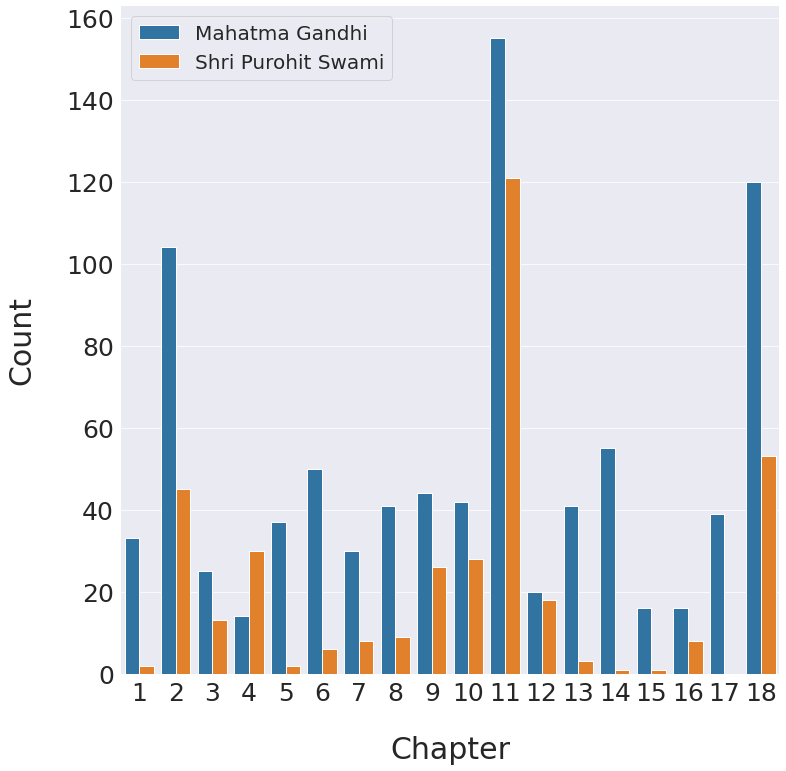}
    \caption{Number of archaic words transformed in translations by Mahatma Gandhi and Shri Purohit Swami.}
    \label{fig:convert}
\end{figure}

In natural language processing applications, there have been studies regarding removal of stop words which can help in removing low level information so that the model can focus on the high level representation \cite{wilbur1992automatic,ghag2015comparative,sarica2021stopwords}. Although it is typical for language models to remove  stop words (eg. "a," "and," "but," "how," "or," and "what"),  in the case of sentiment analysis,  it can eliminate information regarding the context of the sentiment \cite{saif2014stopwords}. Hence, we did not remove stop words for sentiment analysis. Table \ref{tab:preprocessed_texts} presents examples of selected verses from the processed text by Mahatma Gandhi and Shri Purohit Swami after transforming words for preserving semantic information using information given in Table \ref{table:wordconvert}. Figure  \ref{fig:convert} presents the chapter-wise number of occurrences for the word conversions during text processing for the Bhagavad Gita translation by Mahatma Gandhi and Shri Purohit Swami.

\begin{table*}[htbp!]
\small
\centering
\begin{tabular}{lll}
\toprule
Author & Original sentence & Transformed sentence  \\ \hline
\hline
Mahatma Gandhi & \begin{tabular}[c]{@{}l@{}}1. If, O Janardana, thou holdest that the attitude of\\ detachment is superior to action, then why, \\O Keshava, dost thou urge me to dreadful action? \vspace{2 mm}

\end{tabular} & \begin{tabular}[c]{@{}l@{}}If, O Krishna, you hold that the attitude of\\ detachment is superior to action, then why, \\O Krishna, do you urge me to dreadful action? \vspace{2 mm}\end{tabular} \\
\hline 

Mahatma Gandhi & \begin{tabular}[c]{@{}l@{}}27. Whatever thou doest, whatever thou eatest, whatever\\ thou offerest as sacrifice or gift, whatever austerity \\thou dost perform, O kaunteya, dedicate all to Me.\vspace{2 mm}

\end{tabular} & \begin{tabular}[c]{@{}l@{}}Whatever you do, whatever you eat, whatever \\you offer as sacrifice or gift, whatever austerity \\you do perform, O Arjuna, dedicate all to Me.\vspace{2 mm}\end{tabular} \\
\hline 

Shri Purohit Swami & \begin{tabular}[c]{@{}l@{}}7. Yet since with mortal eyes thou canst not see Me, lo! \\I give thee the Divine Sight. See now the glory of \\My Sovereignty.” \vspace{2 mm}

\end{tabular} & \begin{tabular}[c]{@{}l@{}}Yet since with mortal eyes you can not see \\Me, lo! I give you the Divine Sight. See now \\the glory of My Sovereignty.” \vspace{2 mm} \end{tabular} \\
\hline 

Shri Purohit Swami & \begin{tabular}[c]{@{}l@{}}3. Thou art the Primal God, the Ancient, the Supreme \\Abode of this universe, the Knower, the\\ Knowledge and the Final Home. Thou fillest\\ everything. Thy form is infinite. \vspace{2 mm}

\end{tabular} & \begin{tabular}[c]{@{}l@{}}You are the Primal God, the Ancient, the Supreme \\Abode of this universe, the Knower, the Knowledge \\and the Final Home. You fill everything. Your \\form is infinite.  \vspace{2 mm}\end{tabular} \\
\hline 

\end{tabular}
 \caption{Processed text after transforming words for preserving semantic information from Mahatma Gandhi and Shri Purohit Swami.}
    \label{tab:preprocessed_texts}
\end{table*}

\subsection{Sentence Embedding Models}

Sentence BERT (S-BERT)\cite{reimers2019sentence} is a sentence embedding model which improves 
BERT model by reducing computational time to find similar pair of text using Siamese and triplet network structures to derive semantically meaningful sentence embeddings. 
 Although BERT is one of the most successful pre-trained models, it neglects dependency among predicted tokens.    Masked and permuted pre-training \cite{song2020mpnet}  (MPNet) model inherits the features of BERT and and another improved model known as generalized autoregressive pre-training (XLNet) \cite{yang2019xlnet}.  \textit{MPNet-base} encodes sentences into high-dimensional embedding vectors that can be used for various natural language processing tasks. The model takes a variable length English text as an input and gives 768-dimensional output vector. The input sentences of length greater than 384 tokens are truncated. MPNet uses masked and permuted language modeling to model the dependency among predicted tokens and see the position information of the full sentence. The model is also fine-tuned using a contrastive learning objective on a dataset of 1 billion sentence pairs\footnote{\url{https://huggingface.co/sentence-transformers/all-mpnet-base-v2}} for generating the embeddings. Given a sentence from a pair of similar sentences, the model needs to predict the sentence with which it was originally paired from a set of randomly sampled sentences. The sentence embeddings are typically compared using cosine similarity.

 KeyBERT\cite{grootendorst2020keybert} is a keyword extraction technique that leverages BERT embeddings to produce keywords that take into account the semantic aspects of the document. In this method, the input document is embedded using a pre-trained BERT model which embeds  a chunk of text into a fixed-size vector. The keywords and expressions (n-grams) are extracted from the same document using \textit{bag of words} techniques. Each keyword is then embedded into a fixed-size vector with the same model used to embed the document. As the keywords and the document are represented in the same embedding space, KeyBERT computes a cosine similarity between the keyword embeddings and the document embedding. Then, the most similar keywords (with the highest cosine similarity score) are extracted. Furthermore, KeyBERT includes \textit{maximum sum similarity} (MSS) and \textit{maximal marginal relevance} (MMR)  to introduce diversity in the resulting keywords. MMR considers the similarity of keywords/key-phrases with the
    document, along with the similarity of already selected
    keywords and key-phrases. This results in a selection of keywords    that maximizes their within diversity with respect to the document. In our proposed framework, we use KeyBERT along with MMR for  keyword extraction of all the  chapters in different translations of the Bhagavad Gita.

 \subsection{Framework}

\begin{figure*}[htbp!]
\centering
\includegraphics[width=18cm]{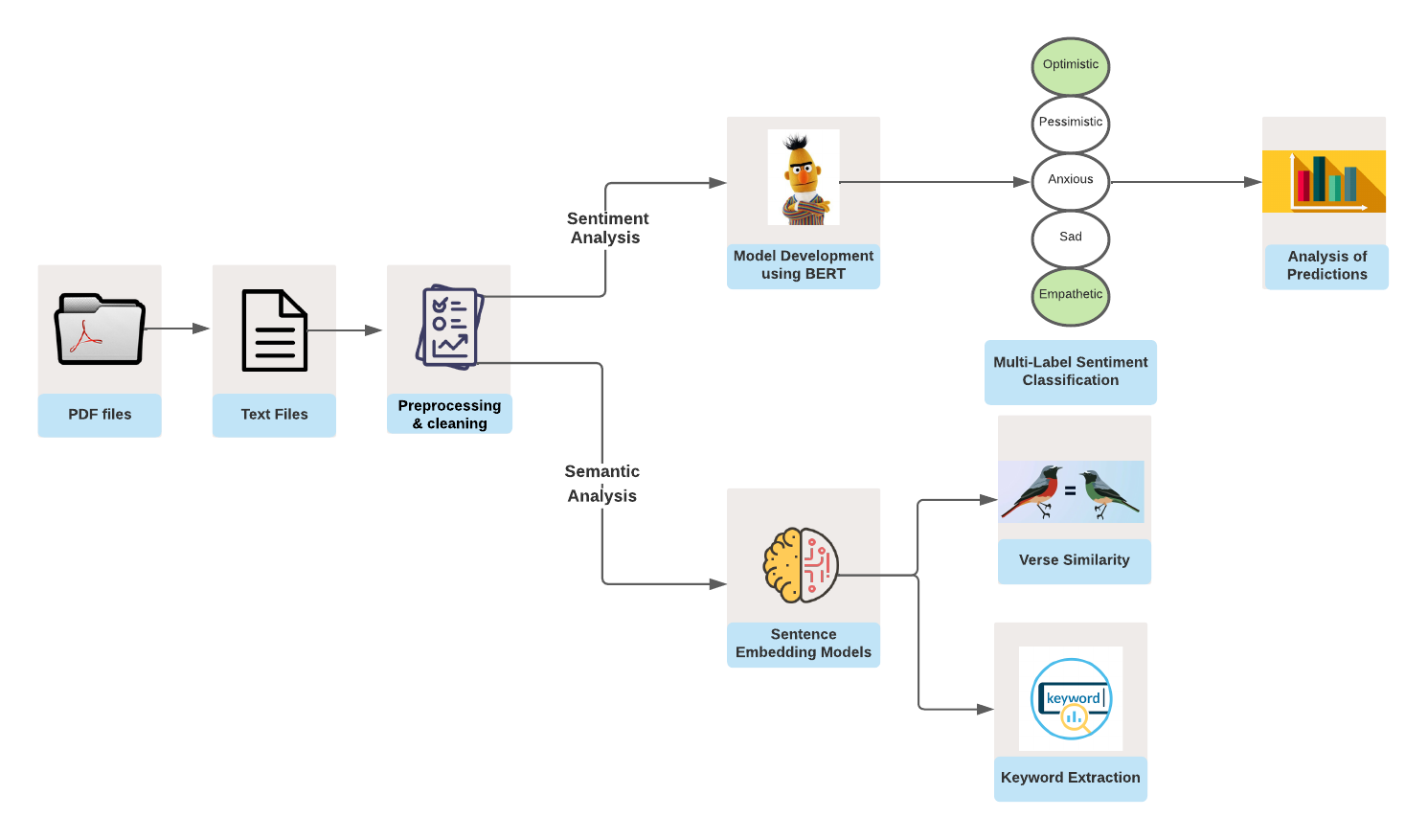}
\caption{Framework for semantic similarity and sentiment analysis for selected translations of the Bhagavad Gita.}
\label{fig:framework}
\end{figure*}

   We present the framework that is used for sentiment and semantic analysis of the Bhagavad Gita.  Our framework analyzes  sentiment and semantic aspects of individual verses from the Bhagavad Gita and also derives insights about the differences in the themes discussed in the texts. As shown in Figure \ref{fig:framework}, we begin by converting the PDF files into text files and  process them as mentioned earlier (Section III -- B). We provide sentiment analysis via  $BERT_{base}$ model by predicting sentiments of verses across the selected translations (texts) of the Bhagavad Gita. We employ multi-label sentiment classification where more than one sentiment can be classified at once, i.e a verse can be both emphatic and optimistic as shown in Figure  \ref{fig:framework}. Although sentiment analysis can be done with model prediction that gives a positive/negative polarity score,   this would not reveal much information about the nature of the sentiment as the positive/negative score can be vague due to different expressions used in translations. Hence, we use the hand-labelled SenWave dataset\cite{yang2020senwave}  which features 11 different sentiments labelled by a group of 50 experts for 10,000 tweets worldwide during COVID-19 pandemic in 2020. The $BERT_{base}$ model is trained using the SenWave dataset \cite{yang2020senwave} for a multi-label sentiment classification task. 
    
 Although sentiment analysis can provide an idea regarding the differences in the translations, it does not provide information about the difference in the semantics (meaning) across the  translations. Hence, semantic analysis \cite{evangelopoulos2012latent} is vital and accompanies sentiment analysis. In the framework, we   provide further analysis of the sentiment predictions by comparing verse-by-verse and chapter-by-chapter statistics of the predicted sentiments across the respective translations.   We also provide semantic analysis via sentence embedding model (MPNet which is based on $BERT_{base}$) and verse-by-verse similarity and perform keyword extraction (based on MPNet) for further analysis of the translations as shown in Figure  \ref{fig:framework}.

 We use the MPNet-base \cite{song2020mpnet} sentence embedding model to encode verses as it produces high quality embeddings.   Since the  embedding is generally in a higher dimension, we reduce the dimension of the encoded vectors to explore the clustering of data in different translations and chapters. We use the   uniform manifold approximation and projection (UMAP) \cite{mcinnes2018umap} algorithm for  dimensionality reduction in order to visualise the high-dimensional vectors. We also extract keywords from the text using KeyBERT to analyze central themes. We note that related methods such as rapid automatic keyword extraction (RAKE) \cite{rose2010automatic}, yet another keyword extractor! (YAKE!)\cite{campos2020yake}, and  term frequency–inverse document frequency (TF-IDF) \cite{salton1983extended} can be used to extract keywords and key-phrases. However, they are based mostly on the statistical properties of a text rather than semantic similarity. Hence, we use KeyBERT as it takes into account the semantic aspects of the text. We filter out the top 10 keywords and sort them in order of their relevance to a given chapter.

\subsection{Model Training} 
 
 We note that although the SenWave dataset is publicly available, permission is required from the authors for usage. In order to ensure reproducibility of the results, and given the constraints, we do not provide the processed dataset. However, we provide trained models in our GitHub repository \footnote{\url{https://github.com/sydney-machine-learning/sentimentanalysis_bhagavadgita}}. The Senwave dataset features 10,000 Tweets that were collected from March to May 2020 labelled with 10 different sentiments plus 1 label which refers to COVID-19 official report. We limit to 10 sentiments and remove the  "official report" in   data processing.  A tweet can feature sentiments  such as  "optimistic", "pessimistic", "anxious", and "thankful". The multi-label sentiment classification task refers to the tweet that can be classified as "optimistic" and "anxious" at once.   We train the $BERT_{base}$ model on SenWave dataset  by pre-processing the tweets using our previous work \cite{chandra2021covid}.  
   
    The pre-trained BERT-base uncased model is further tuned on the  SenWave dataset \cite{yang2020senwave} with training and test dataset created in a 90/10 \% split on a dataset of 10,000 tweets using a batch size of 1 using 4 training epochs. We use the default hyper-parameters for the  model and implement regularisation using a dropout\cite{srivastava2014dropout} layer with dropout probability of 0.3 to avoid over-fitting. Finally, we add a linear activation layer with 10 outputs to predict 10 sentiments. Moreover, we use the adaptive moment estimation (Adam) gradient  \cite{kingma2017adam} optimizer  for training with a learning rate of 1e-05, and  binary cross entropy loss function for the multi-label classification task.  
    
 We encode verses using the MPNet-base model and compute the verse by verse semantic similarity. We also use the MPNet-base model for extraction of keywords (using KeyBERT) from all chapters. However, given the constraint in the MPNet-base model that number of tokens should not exceed 384, it would not be possible to encode large chapters directly. Hence, we propose a method to overcome this limitation by breaking each chapter into paragraphs of 15 verses each. We include 3 verses from the previous paragraph into the current paragraph to retain some context and maintain continuity. For example, in the first paragraph, verses 1-15 are included, and in the second paragraph verses 13-27, then 25-39, and so on.  We keep the top 20 keywords because keywords that have a lower similarity score in the original paragraph may be more relevant when the entire chapter is considered. 
 
 Next, we extract the keywords for all paragraphs $i$, with 20 candidate keywords of paragraph $j$, such that $i$ $\neq$ $j$. For each keyword, we add up its cosine similarity score for different paragraphs. Finally, we obtain the top 10 keywords having the highest cumulative scores. The key idea here is that if a term is a keyword in a certain paragraph, it also needs to be a keyword in other paragraphs for it to qualify as a keyword for the entire chapter. We use  maximal marginal relevance (MMR) with a diversity value of 0.5 to prevent the selection of similar meaning keywords.

  \section{Results}

\subsection{Data Analysis}

\begin{figure}[htbp!]
    \subfigure[Eknath Easwaran ]{\label{fig:easwaran_bigrams}\includegraphics[width = 0.99\linewidth]{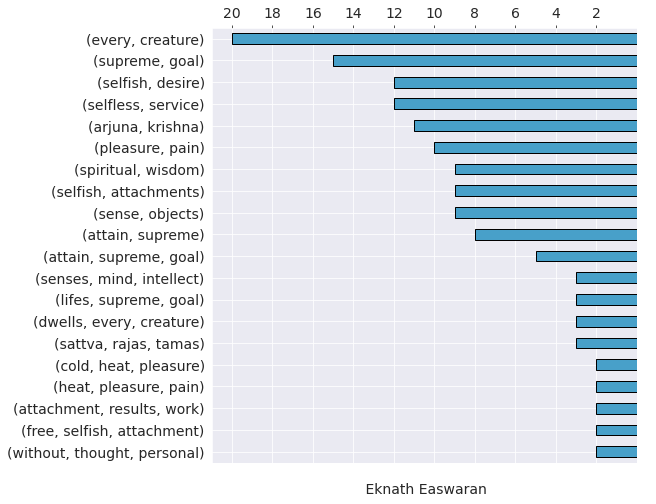}}
    
    \subfigure[Shri Purohit Swami ]{\label{fig:purohit_bigrams}\includegraphics[width = 0.99\linewidth]{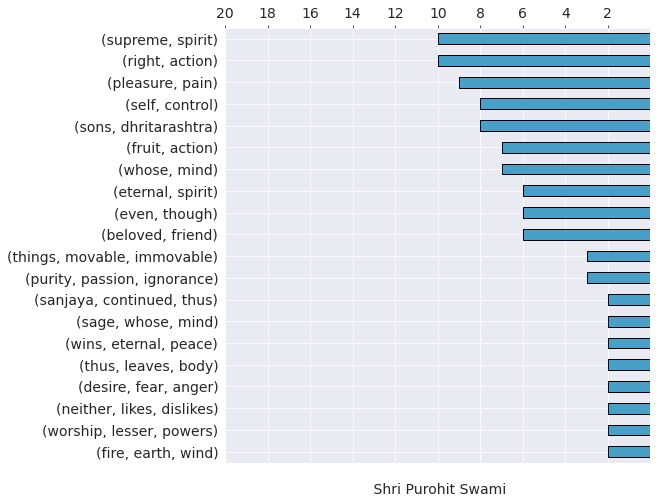}}
    
    \subfigure[Mahatma Gandhi]{\label{fig:mahatma_gandhi_bigrams}\includegraphics[width=0.99\linewidth]{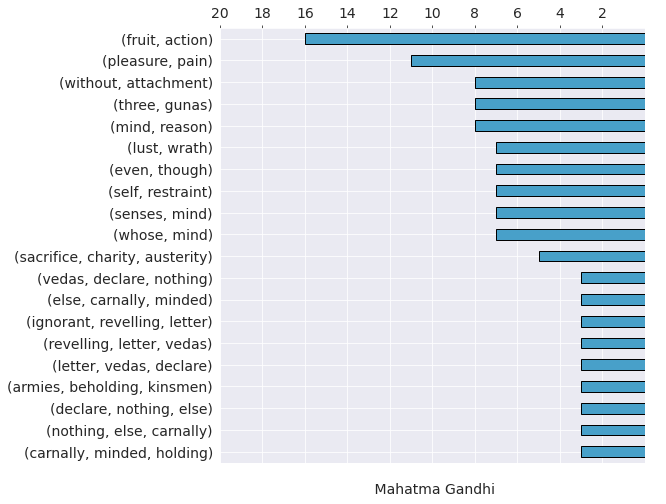}}
    
    \caption{Top 10 bigrams and trigrams from processed verses of all translations.}
    \label{complete_text_bigrams}
\end{figure}

We first present results with basic data analysis for visualisation of the datasets.  Figure \ref{complete_text_bigrams}   presents the bigrams and trigrams obtained for the respective translations of the Bhagavad Gita. We removed the stop words in order to obtain the bigrams and trigrams. Throughout the Bhagavad Gita, the concept of three 'gunas' or attributes has been widely discussed. Dominance of the 'sattvic' attribute leads to liberation, that of 'rajasa' attribute leads to rebirth as humans and 'tamas' means moving to a lower form of life. Hence, the bigram 'three gunas' has been aptly extracted in Gandhi's version (Figure \ref{fig:mahatma_gandhi_bigrams}). Also, the three major human attributes (gunas) i.e. sattva, rajas, tamas in Easwaran's translation roughly translate to the attributes purity, passion and ignorance as described in Shri Purohit Swami's translation (Figure \ref{fig:purohit_bigrams}).
  
  We observe that  the bigram, \textit{pleasure and pain}  appears in all translations. This gives the message the a person who is fit for liberation is unaffected by worldly (material) desires and   at complete peace with themselves,  unaffected by either pleasure or pain. Since the bigrams and trigrams merely count the occurrence of a set of words, the central themes that are emphasised will become evident afterwards. While all three texts discuss the idea of 'supreme spirit' or the \textit{Atman}; however, the path taken to realize it differs. Eknath Easwaran stresses the importance of \textit{selfless service} devoid of \textit{selfish attachments} and \textit{desires} (Figure \ref{complete_text_bigrams}). Shri Purohit Swami advocates the path of \textit{right} (righteous) \textit{action} by observing \textit{self control}. Interestingly,  Chapter 3 of Eknath Easwaran's translation is titled 'Selfless Service' while Shri Swami Purohit has titled it as 'Karma Yoga - The Path of Action'. Mahatma Gandhi observes that the mind would have to be completely \textit{self restrained} to be able to renounce the \textit{attachment to fruits of action}. The essence of Mahatma Gandhi's interpretation is beautifully captured in verse 47 of Chapter 2 - \textit{"Action alone is thy province, never the fruits thereof; let not thy motive be the fruit of action, nor shouldst thou desire to avoid action." }
  
  Furthermore, we notice that the top three bigrams and trigrams (Figure 3) across the three different translations are almost entirely different. It seems that the different translators have used different words to refer to similar concepts, for example, Eknath Easwaran's translation features [every, creature] and [supreme, goal] as top bigrams whereas Shri Purohit Swami's translation gives [supreme, spirit] and  [right, action]. Moreover, Mahatama Gandhi's translation features [fruit, action] and  [pleasure, pain] which is also present in Eknath Easwaran's translation. Hence, a word by word comparison  using bigrams and trigrams shows  a major difference in the respective translations.

Table \ref{table:chapter_names}  lists all the chapter titles of the three translations. In Mahatma Gandhi's translation, chapters are referred to as 'discourses'.   We note the difference in the way the different translators interpreted the title of the respective chapters. Eknath Easwaran, for example, translated Chapter 2 heading as "Self-Realization" while Shri Purohit Swami translated it as "The Philosophy of Discrimination". 

\begin{table*}[hbp!]
\centering 
\begin{tabular}{| c | c | c | c |} 
\hline\hline 
Chapter No. & Eknath Easwaran & Shri Swami Purohit & Mahatma Gandhi\\ [1ex] 
\hline 
1 & The War Within & The Despondency of Arjuna  &  - \\ [0.5ex] 
2 & Self-Realization  &  The Philosophy of Discrimination & - \\ [0.5ex]
3 & Selfless Service   & Karma-Yoga  - The Path of Action &  - \\ [0.5ex]
4 & Wisdom in Action  & Dnyana-Yoga  -  The Path of Wisdom & - \\ [0.5ex]
5 & Renounce and Rejoice & The Renunciation of Action  &  - \\ [0.5ex]
6 & The Practice of Meditation & Self-Control  &  - \\ [0.5ex]
7 & Wisdom from Realization & Knowledge and Experience & - \\ [0.5ex]
8 & The Eternal Godhead & Life Everlasting & - \\ [0.5ex]
9 & The Royal Path & The Science of Sciences and the Mystery of Mysteries  & - \\ [0.5ex]
10 & Divine Splendor & The Divine Manifestations & - \\ [0.5ex]
11 & The Cosmic Vision & The Cosmic Vision & - \\ [1ex]
12 & The Way of Love & Bhakti-Yoga  -  The Path of Love & - \\ [0.5ex]
13 & The Field and the Knower & Spirit and Matter & - \\ [0.5ex]
14 & The Forces of Evolution & The Three Qualities & - \\ [0.5ex]
15 & The Supreme Self & The Lord-God & - \\ [0.5ex]
16 & Two Paths & Divine and Demonic Civilization & - \\ [0.5ex]
17 & The Power of Faith & The Threefold Faith & - \\ [0.5ex]
18 & Freedom and Renunciation & The Spirit of Renunciation & - \\ [1ex]
\hline 
\end{tabular}
\caption{Names of Chapters in the Bhagavad Gita as given in the  different translations.}
\label{table:chapter_names}
\end{table*}

\subsection{Sentiment Analysis}

Next, we use  the BERT model for verse-by-verse sentiment analysis of the respective Bhagavad Gita translations. We first present the bigrams and the trigrams of the top ten optimistic and pessimistic sentiments for the entire text for the respective translations. 

The sentiment  analysis shows that  [pleasure, pain] is an optimistic bigram in all three translations (Figure \ref{fig:optimistic_vs_pessimistic_easwaran}, Figure \ref{fig:optimistic_vs_pessimistic_gandhi}, Figure \ref{fig:optimistic_vs_pessimistic_purohit}). However, this has been used in the context that one should have complete control over one's mind, which is one of the central themes in the Bhagavad Gita. According to the Bhagavad Gita, the mind should not be clouded by either pleasure or pain, by heat or cold, or by other worldly influences (Chapter 2, verse 15). Only such a person is fit for immortality (moksha) to escape the cycle of birth and death  (Chapter 2, verse 15).  Lord Krishna uses the contradicting concepts of pleasure and pain to demonstrate this theme about the human condition \cite{bednarik2011human} in order to convince Arjuna to pick his bow up and rise above worldly emotions.

Moreover, in terms of leading optimistic sentiments predicted by the model, according to the bigrams -  we find [supreme, goal] and [every, creature] and the leading optimistic sentiment bigrams for the Eknath Easwaran's translation (Figure \ref{fig:optimistic_vs_pessimistic_easwaran}), whereas Shri Purohit Swami's translation gives [supreme, spirit] and [self, control] (Figure \ref{fig:optimistic_vs_pessimistic_purohit}). Mahatma Gandhi's translation gives [pleasure, pain] and [without, attachment] (Figure \ref{fig:optimistic_vs_pessimistic_gandhi}). We notice that [pleasure, pain] is listed in the top five optimistic bigrams across the respective translation which shows that although the wording used in the translations are different, the sentiments they express are similar. Interestingly, we notice that [every, creature] (Figure \ref{fig:optimistic_vs_pessimistic_easwaran}) and [fruit, action] (Figure \ref{fig:optimistic_vs_pessimistic_purohit}) is listed in both pessimistic and optimistic sentiments in the respective translations. A major reason could be the context in which these terms appear in the respective verses in the text.

\begin{figure}[!htb]
    \centering
    \includegraphics[scale=0.25]{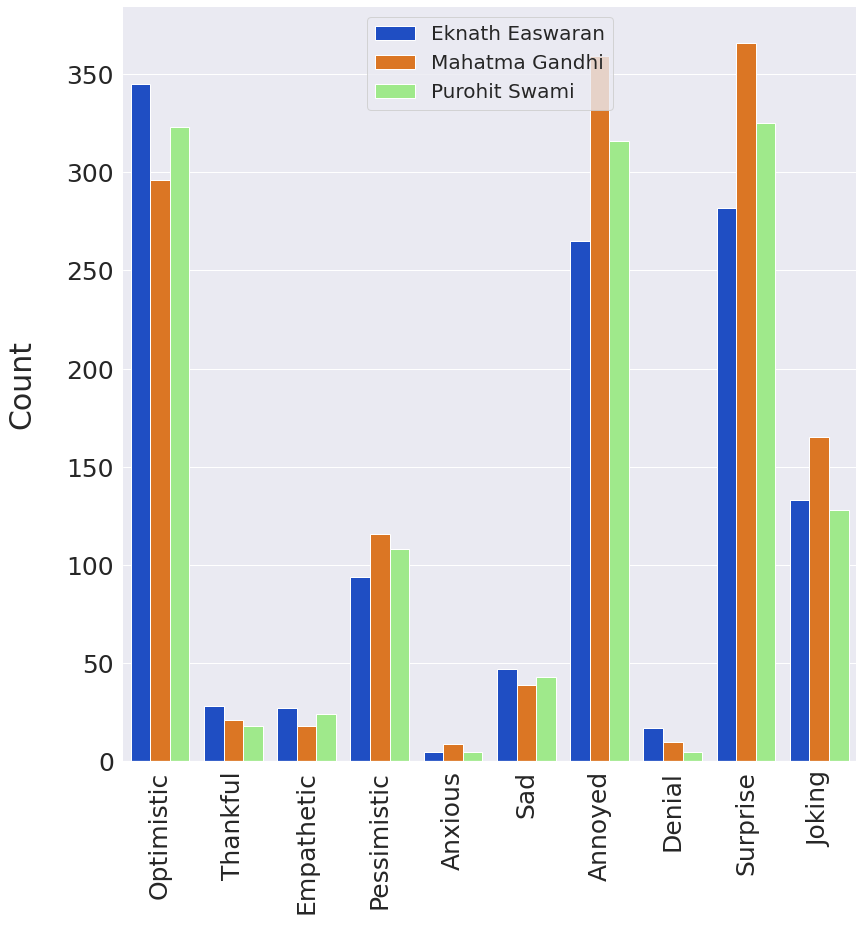}
    \caption{Text-wise sentiment analysis.}
    \label{fig:barplot_all_chapters}
\end{figure}

\begin{figure*}[hbp!]
    \centering
    \includegraphics[width=\textwidth]{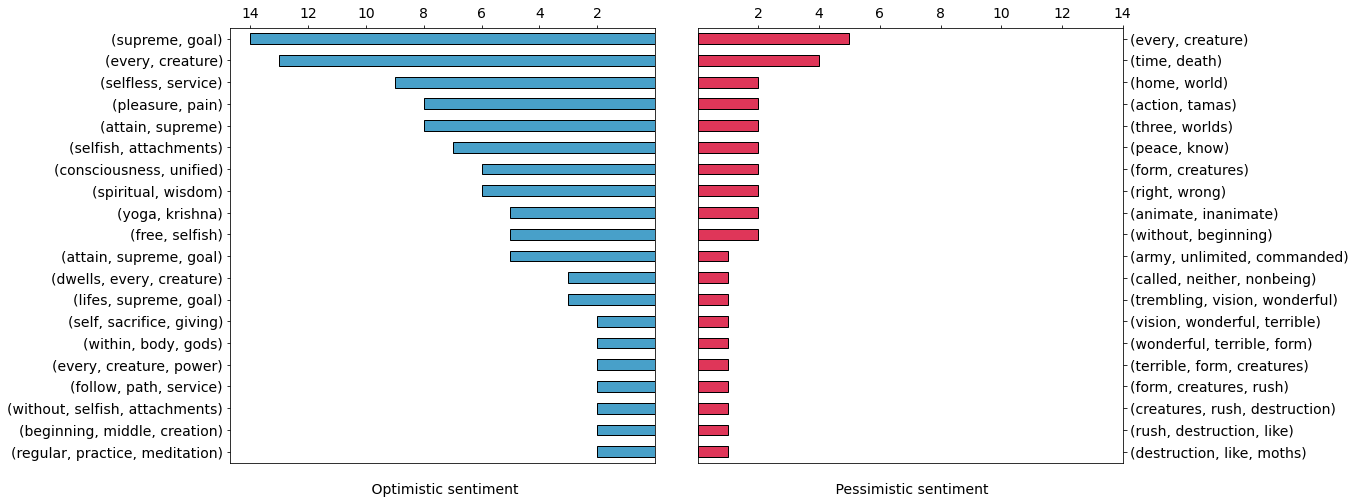}
    \caption{Top 10 optimistic and pessimistic bigrams and trigrams from processed verses of Eknath Easwaran's translation.}
    \label{fig:optimistic_vs_pessimistic_easwaran}
\end{figure*}

\begin{figure*}[hbp!]
    \centering
    \includegraphics[width=\textwidth]{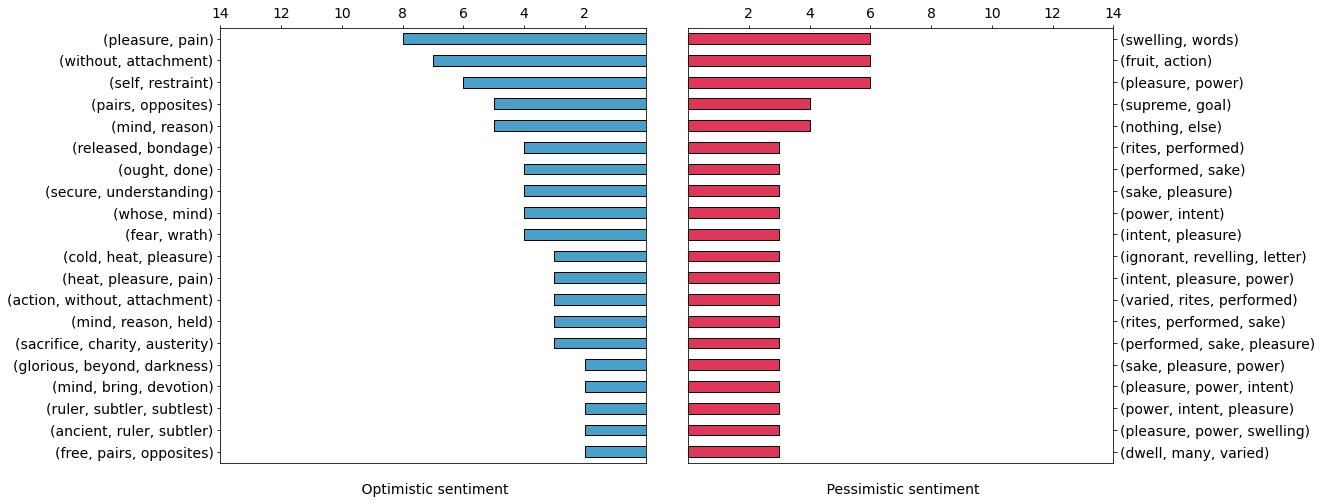}
    \caption{Top 10 optimistic and pessimistic bigrams and trigrams from processed verses of Mahatma Gandhi's translation.}
    \label{fig:optimistic_vs_pessimistic_gandhi}
\end{figure*}

\begin{figure*}[hbp!]
    \centering
    \includegraphics[width=\textwidth]{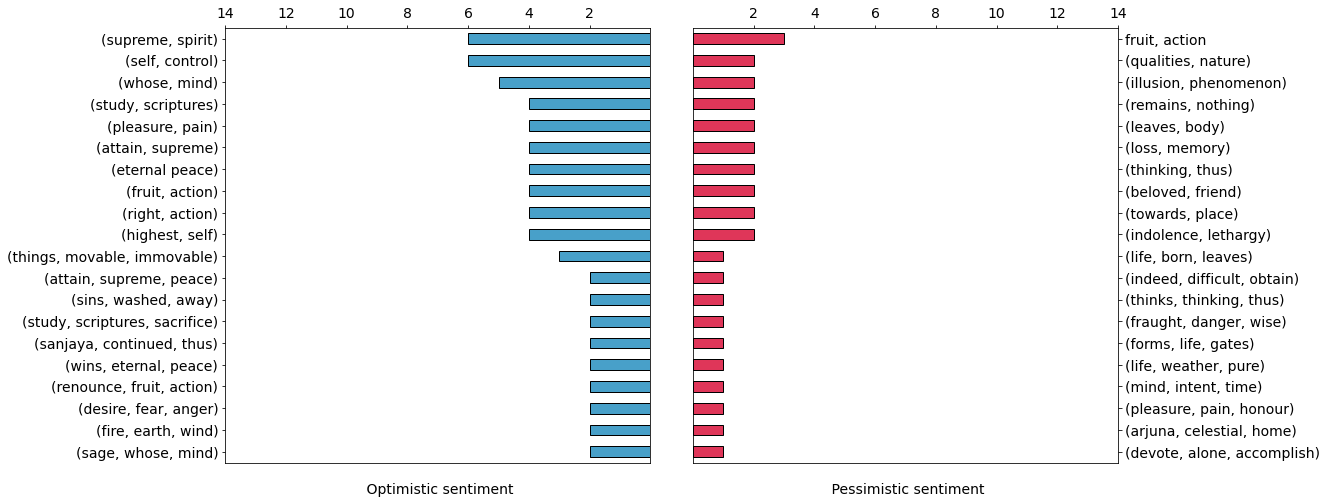}
    \caption{Top 10 optimistic and pessimistic Bi-grams and Tri-grams from processed verses for Shri Swami Purohit's translation}
    \label{fig:optimistic_vs_pessimistic_purohit}
\end{figure*}

\begin{figure*}[!htb]
  \centering
\subfigure[Chapter 1: The War Within]{\label{fig:1}\includegraphics[width=.32\linewidth]{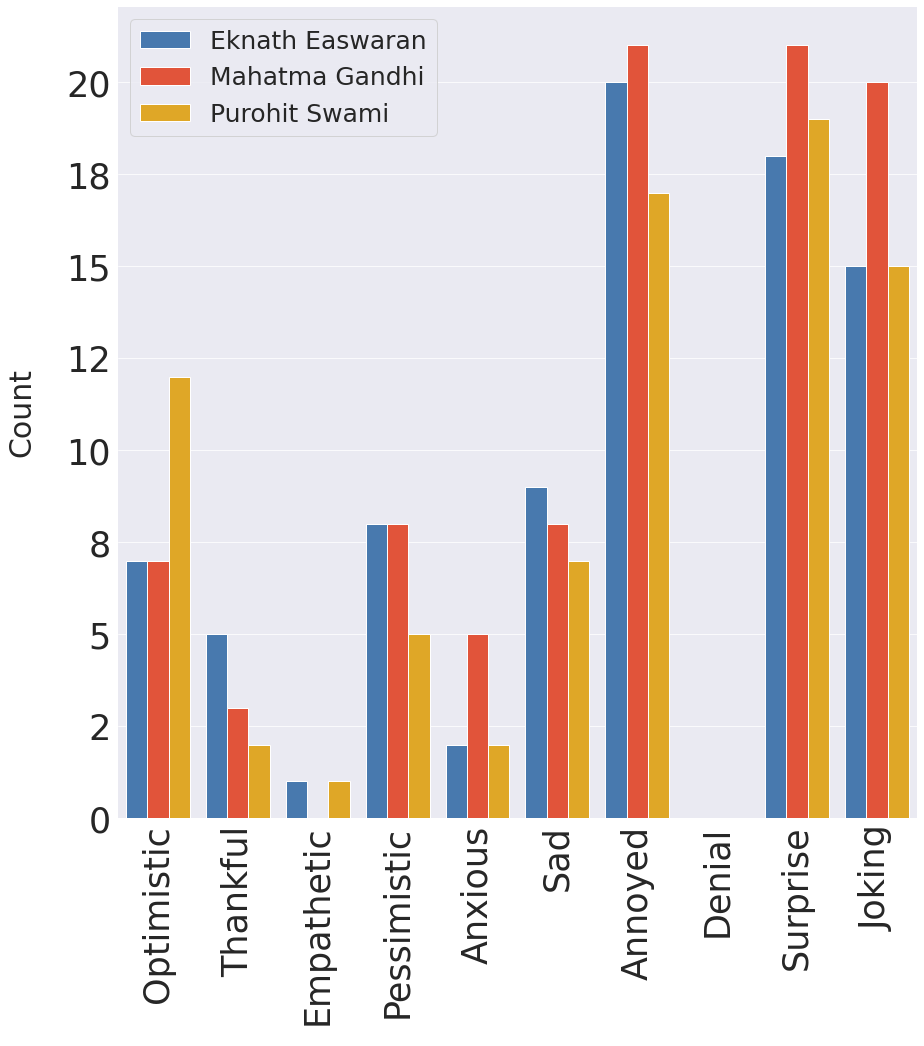}}
\subfigure[Chapter 2: Self-Realization ]{\label{fig:2}\includegraphics[width=.32\linewidth]{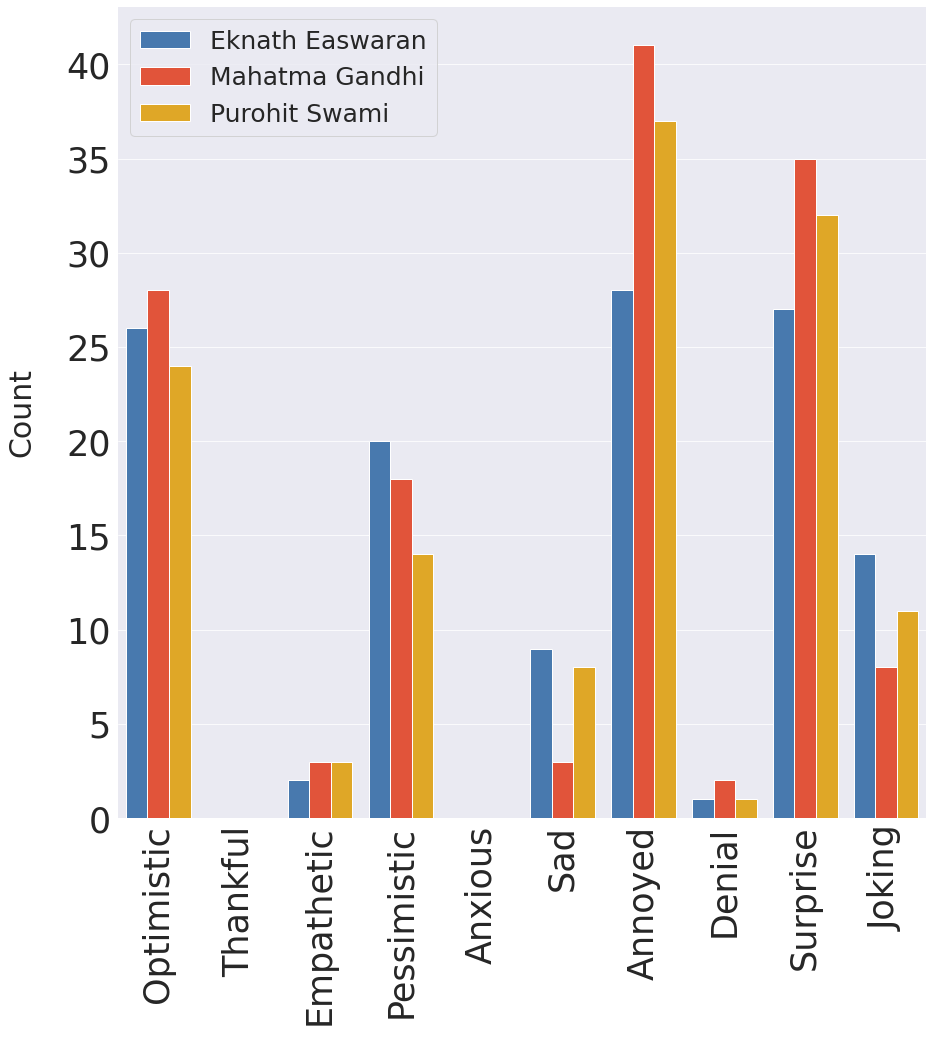}}
\subfigure[Chapter 3: Selfless Service]{\label{fig:c}\includegraphics[width=.32\linewidth]{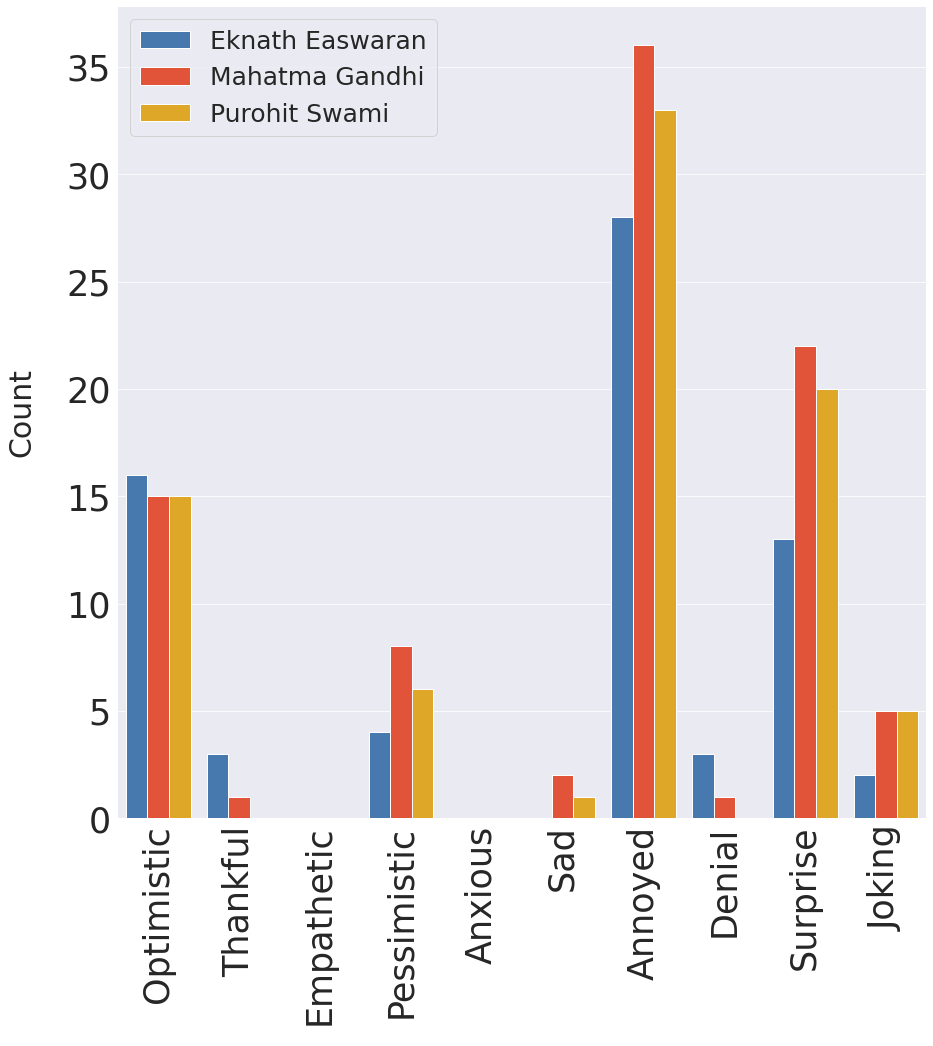}}
\subfigure[Chapter 4: Wisdom in Action ]{\label{fig:d}\includegraphics[width=.32\linewidth]{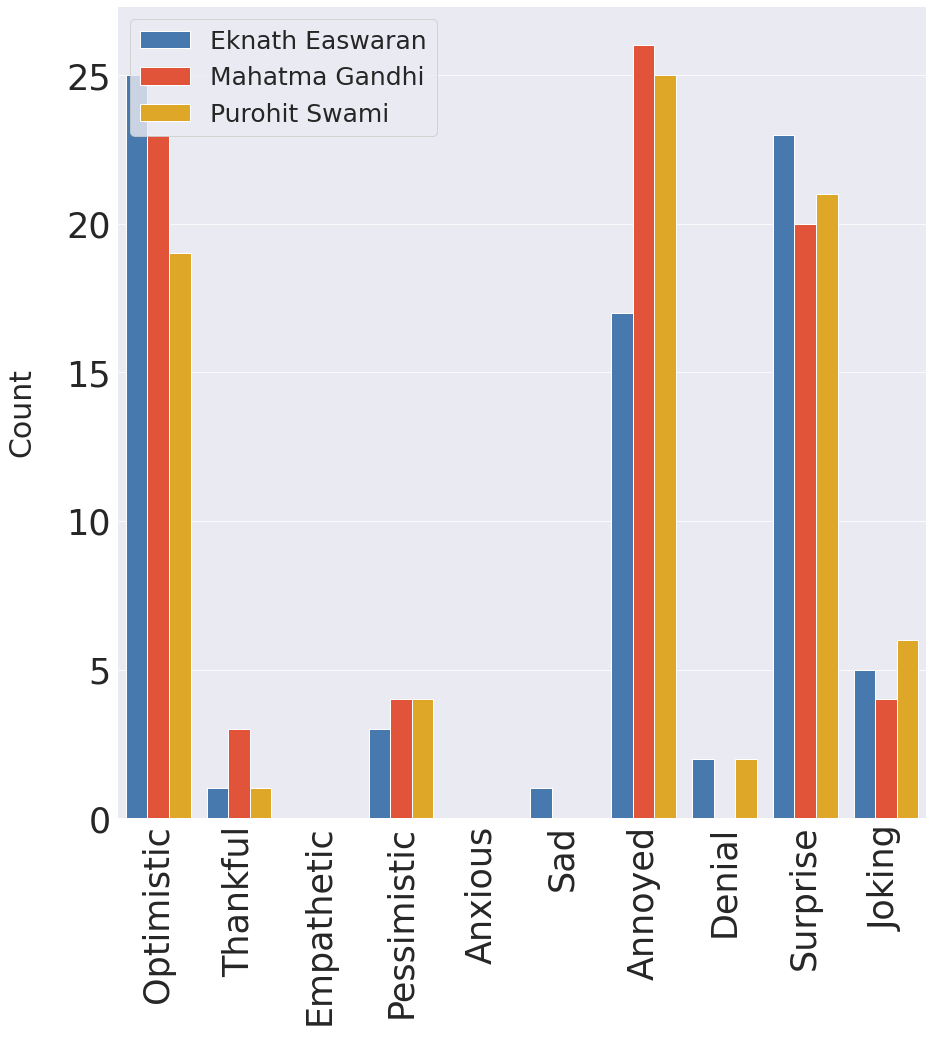}}
\subfigure[Chapter 5: Renounce and Rejoice]{\label{fig:e}\includegraphics[width=.32\linewidth]{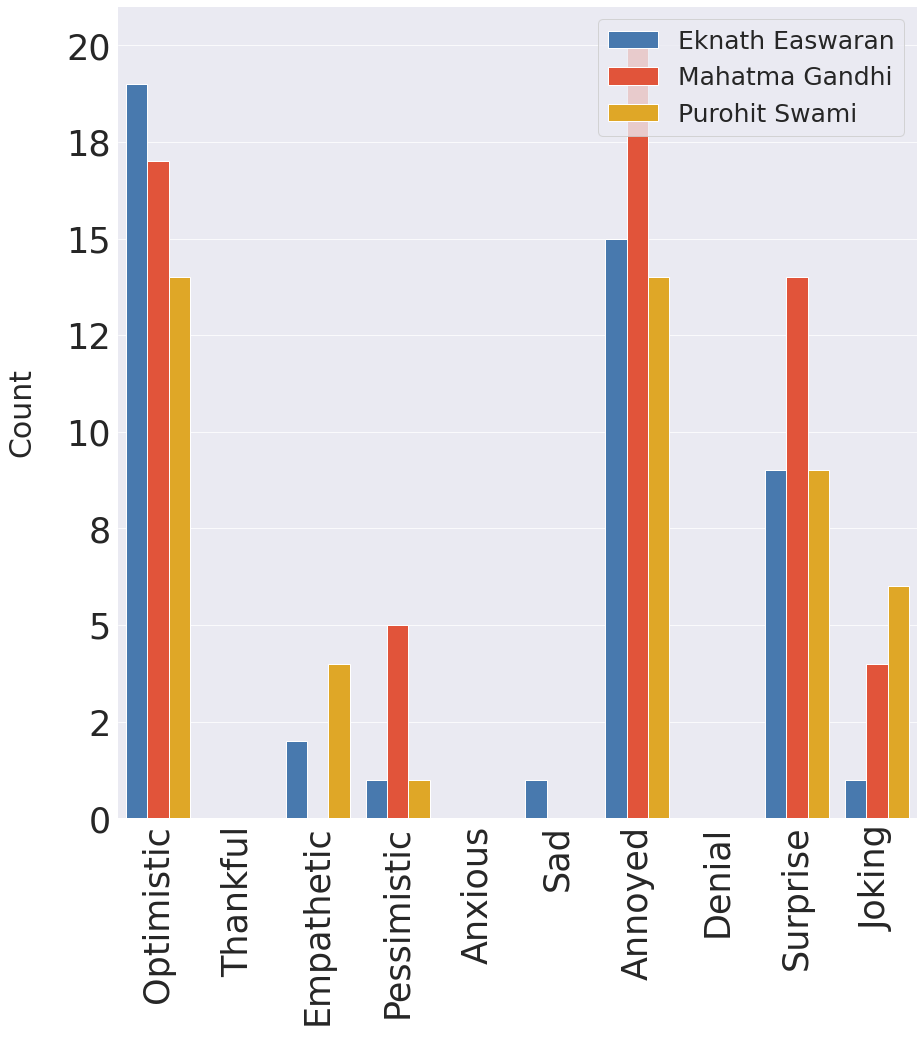}}
\subfigure[Chapter 6: The Practice of Meditation ]{\label{fig:f}\includegraphics[width=.32\linewidth]{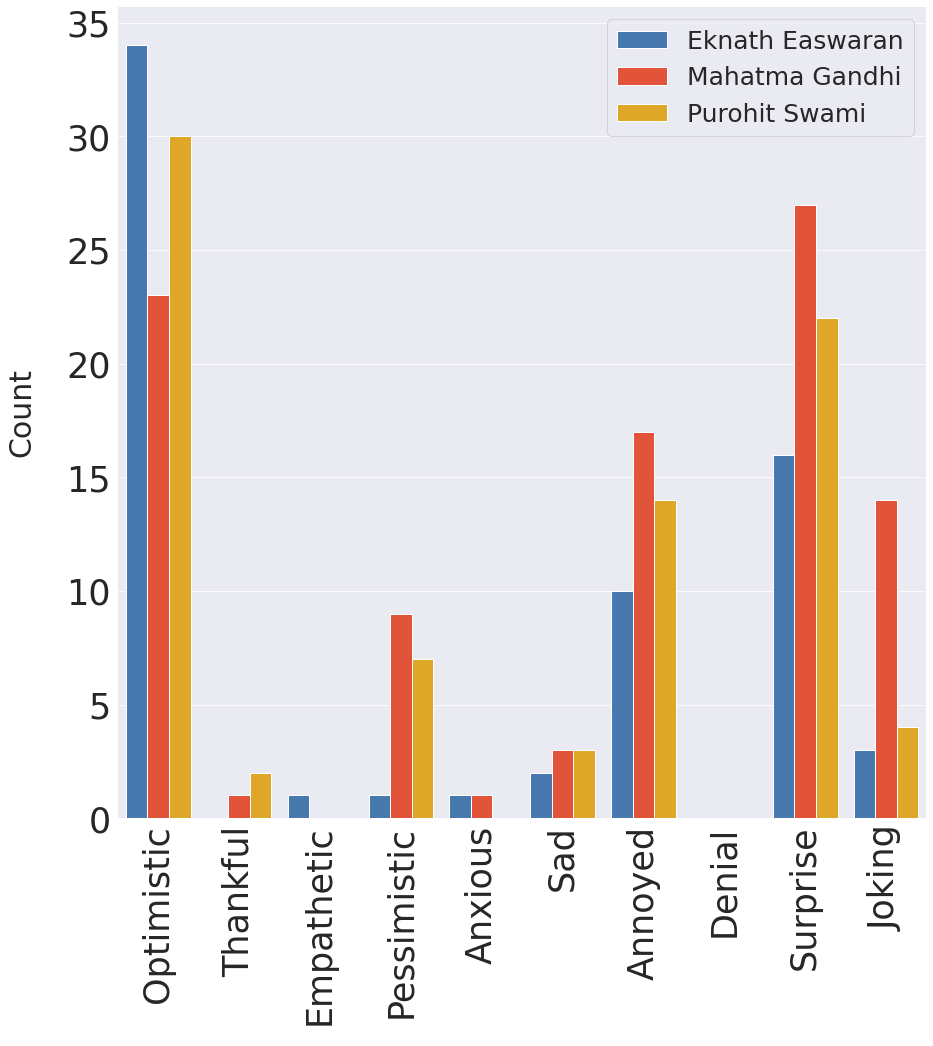}}
\subfigure[Chapter 7: Wisdom and Realization]{\label{fig:g}\includegraphics[width=.32\linewidth]{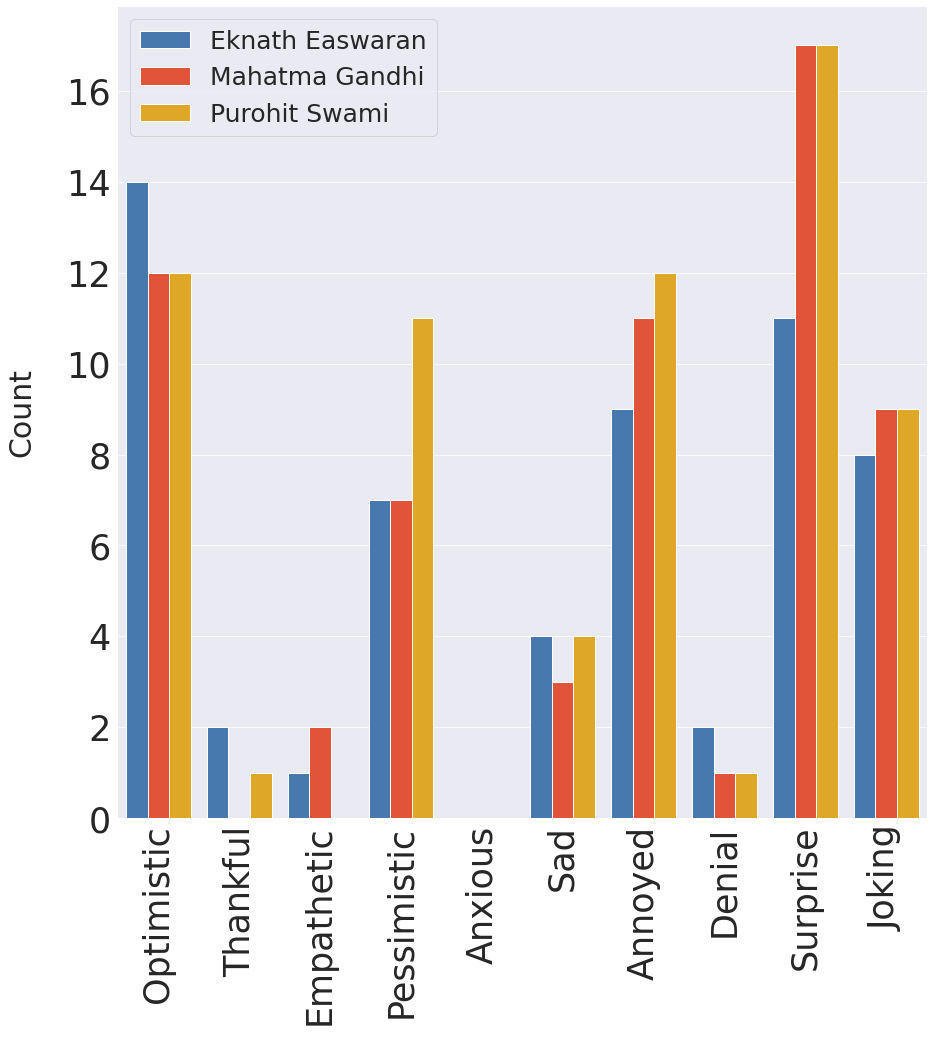}}
\subfigure[Chapter 8: The Eternal Godhead]{\label{fig:h}\includegraphics[width=.32\linewidth]{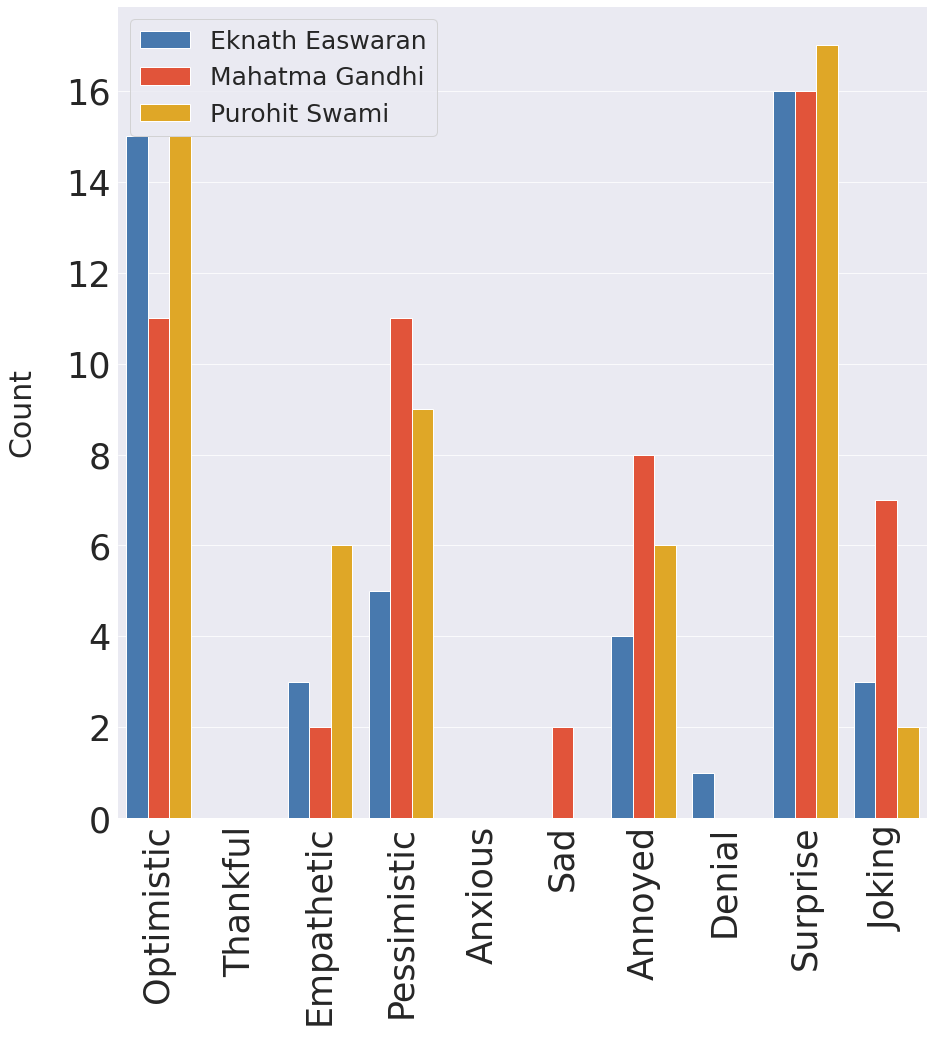}}
\subfigure[Chapter 9: The Royal Path]{\label{fig:i} \includegraphics[width=.32\linewidth]{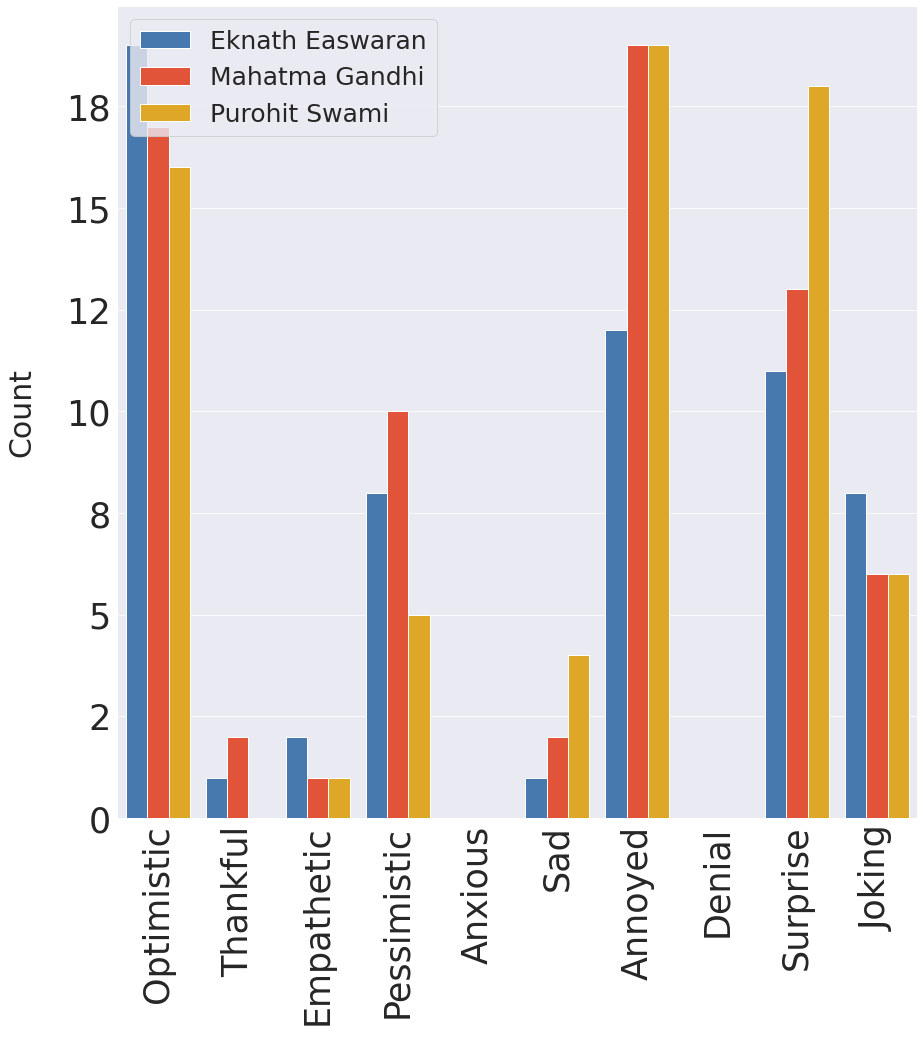}}
\caption{Chapter-wise sentiment analysis (Chapters 1 - 9.)}
\label{fig:chaptersent}
\end{figure*}

\begin{figure*}[!htb]
    \centering
    
\subfigure[Chapter 10: Divine Splendor]{\label{fig:s}\includegraphics[width=.32\linewidth]{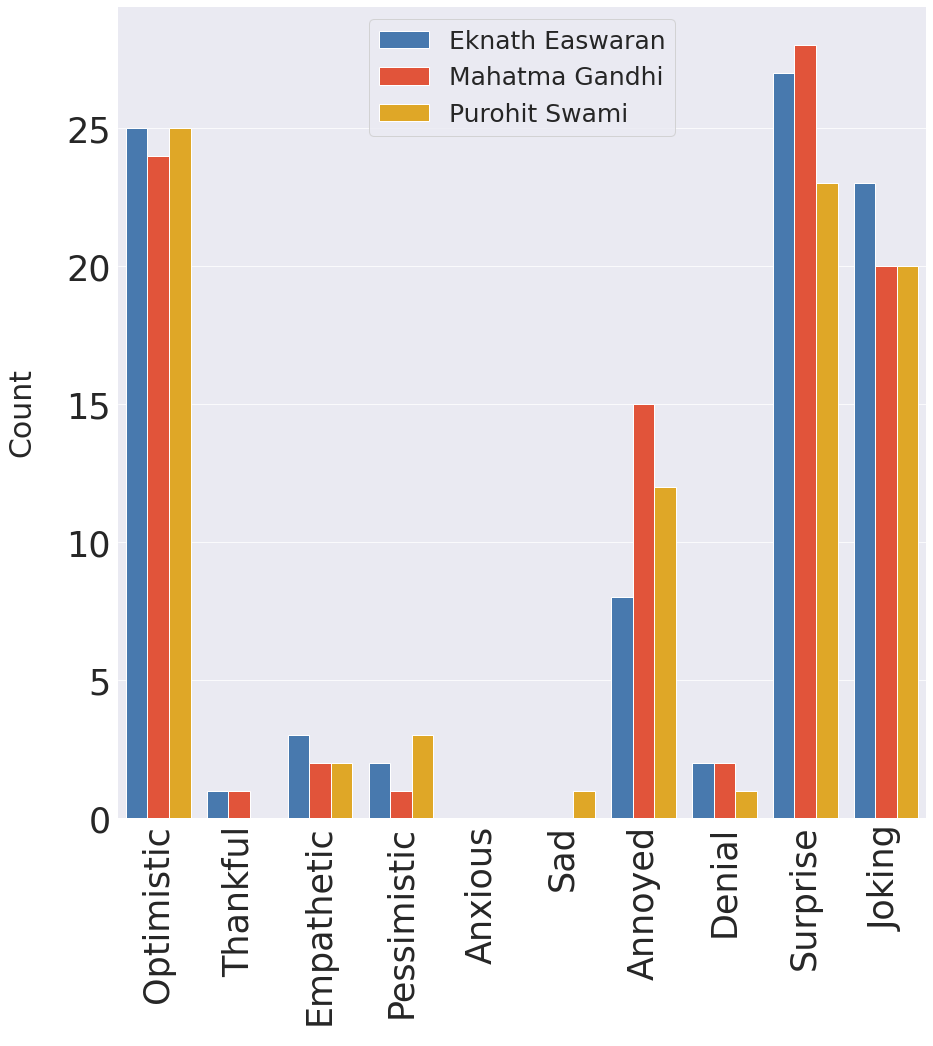}}
\subfigure[Chapter 11: The Cosmic Vision]{\label{fig:j}\includegraphics[width=.32\linewidth]{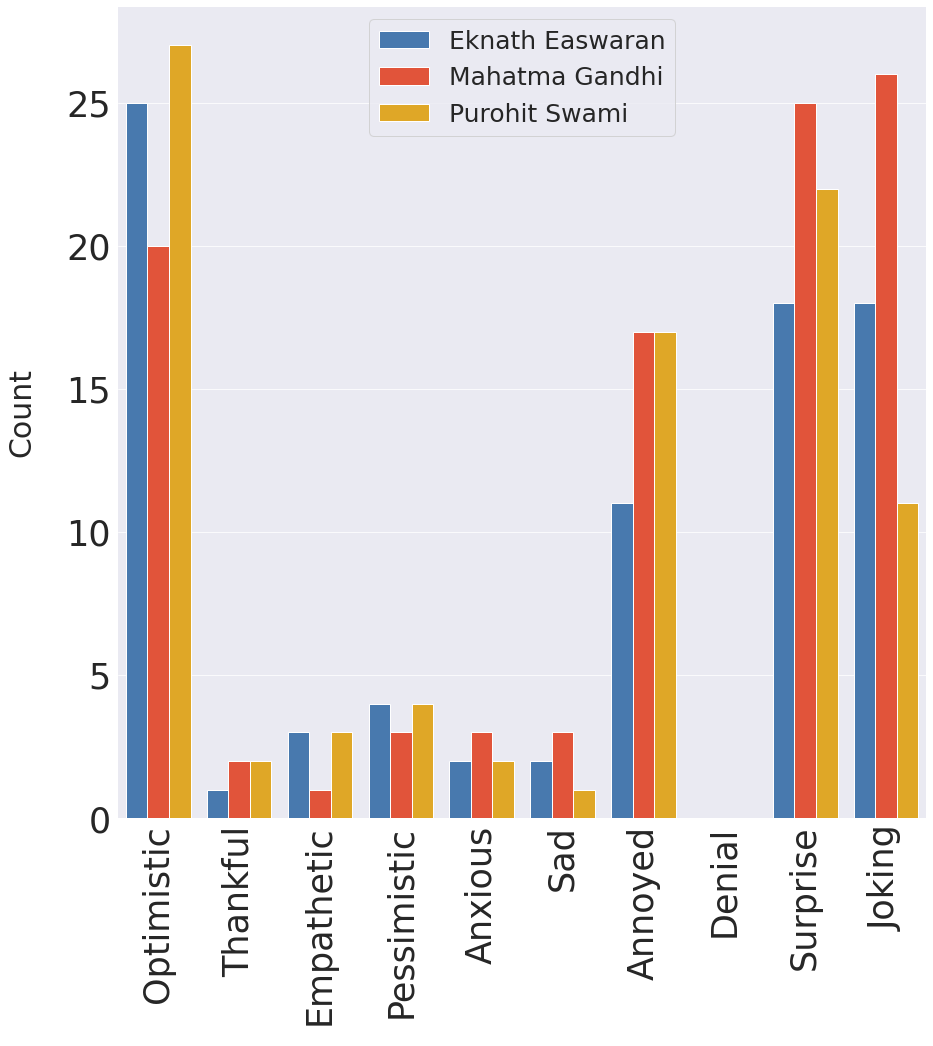}}
\subfigure[Chapter 12: The Way of Love]{\label{fig:k}\includegraphics[width=.32\linewidth]{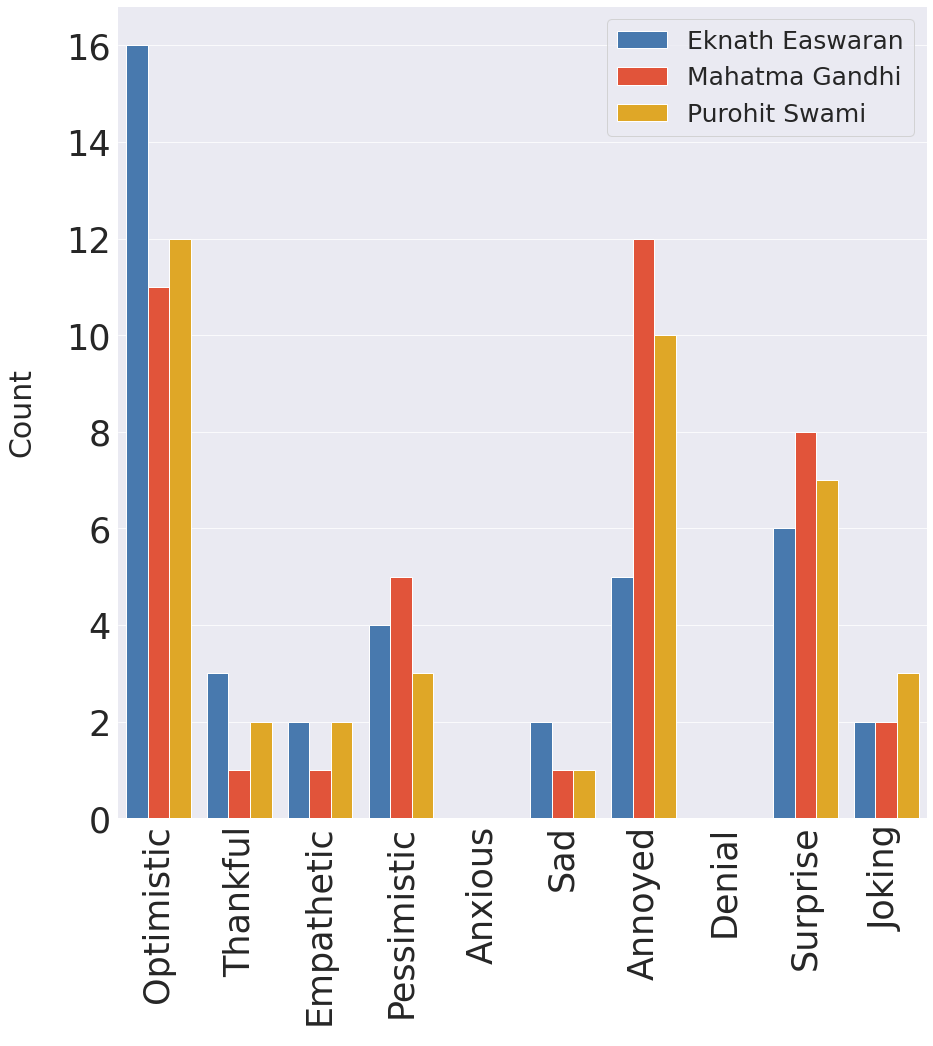}}
\subfigure[Chapter 13: The Field and the Knower]{\label{fig:l}\includegraphics[width=.32\linewidth]{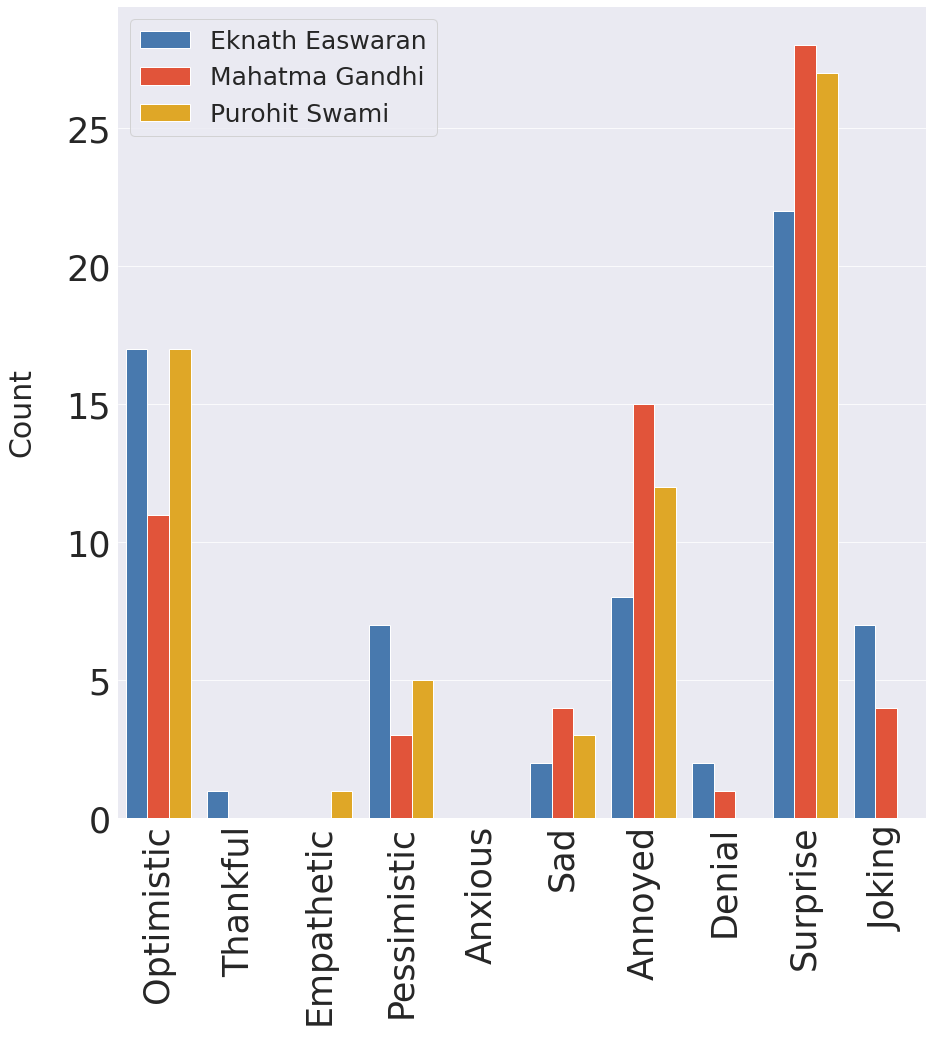}}
\subfigure[Chapter 14: The Forces of Evolution]{\label{fig:m}\includegraphics[width=.32\linewidth]{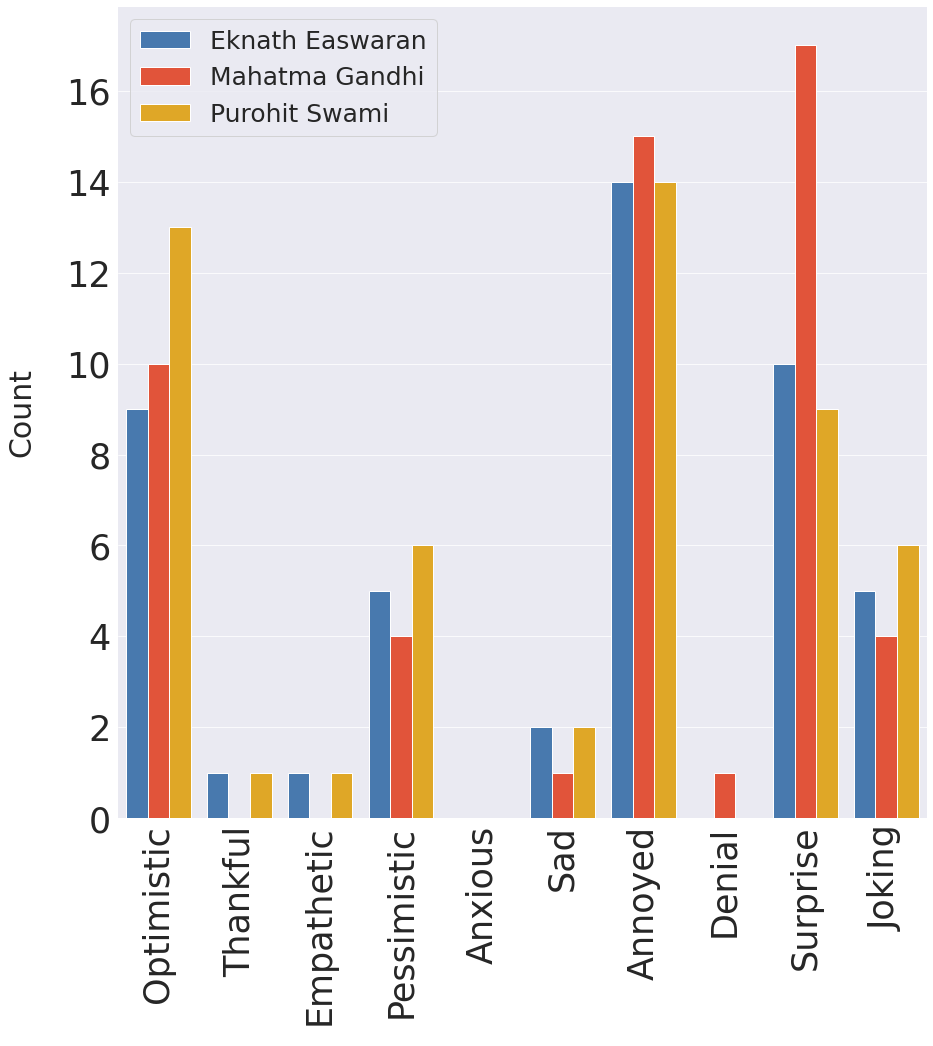}}
\subfigure[Chapter 15: The Supreme Self]{\label{fig:n}\includegraphics[width=.32\linewidth]{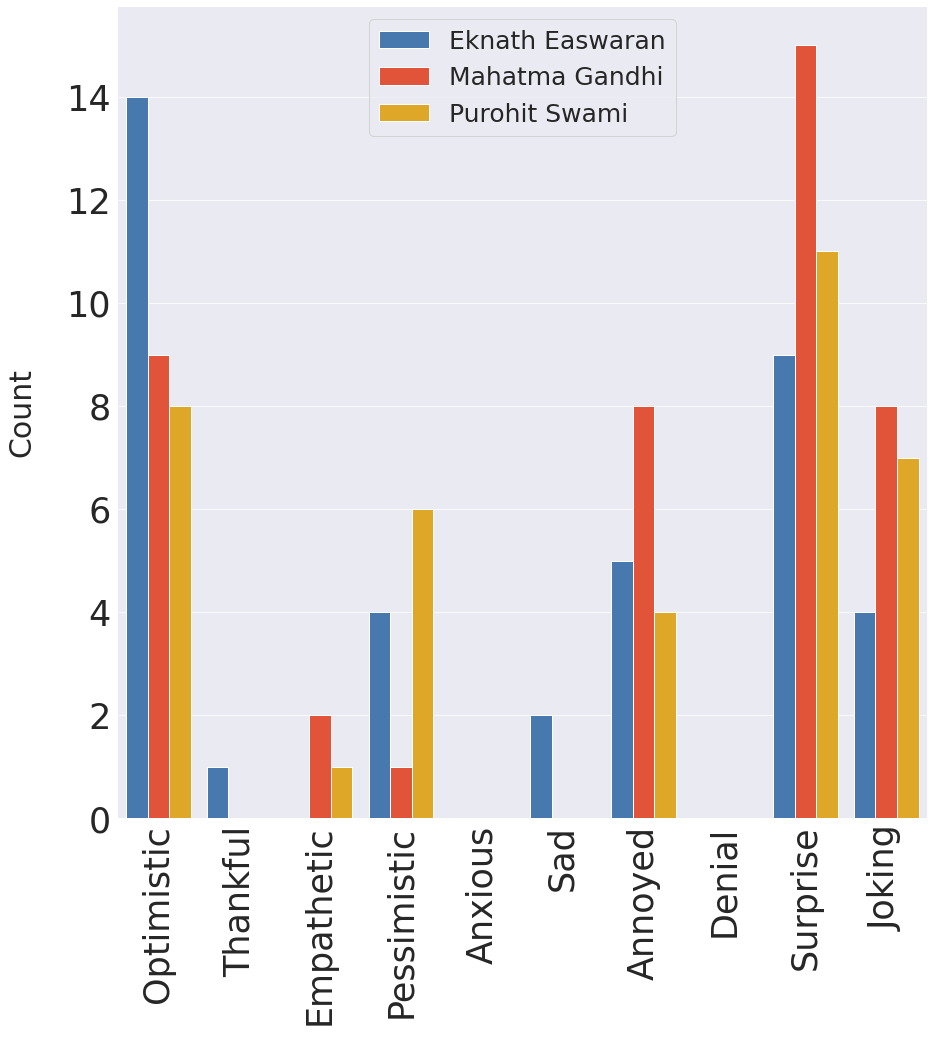}}
\subfigure[Chapter 16: Two Paths]{\label{fig:o}\includegraphics[width=.32\linewidth]{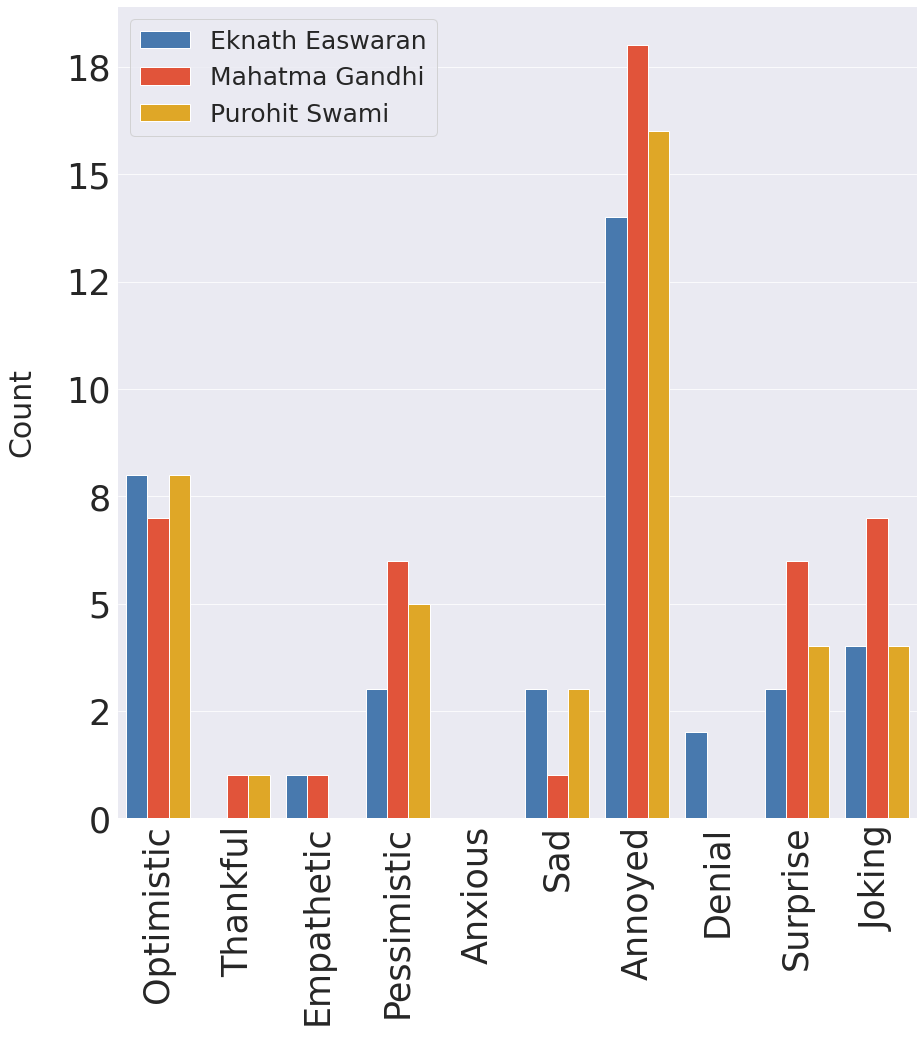}}
\subfigure[Chapter 17: The Power of Faith]{\label{fig:p}\includegraphics[width=.32\linewidth]{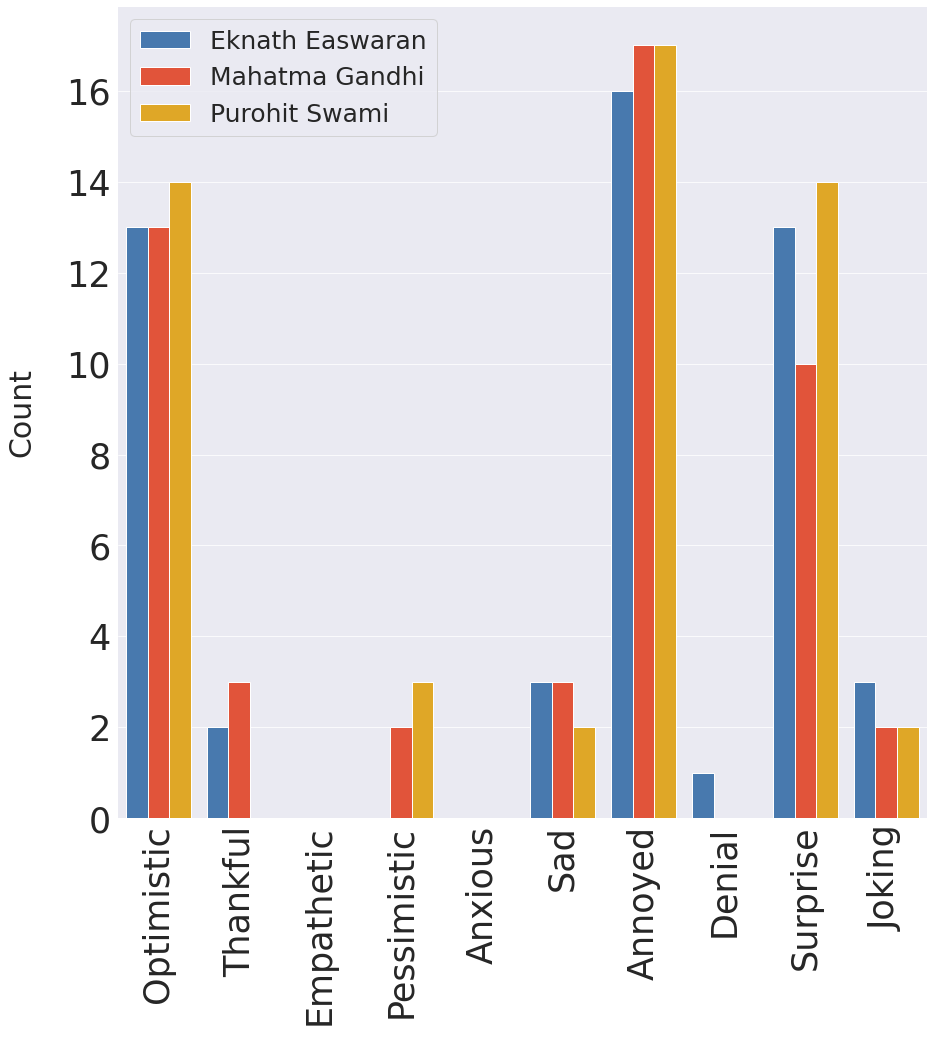}}
\subfigure[Chapter 18: Freedom and Renunciation]{\label{fig:q}\includegraphics[width=.32\linewidth]{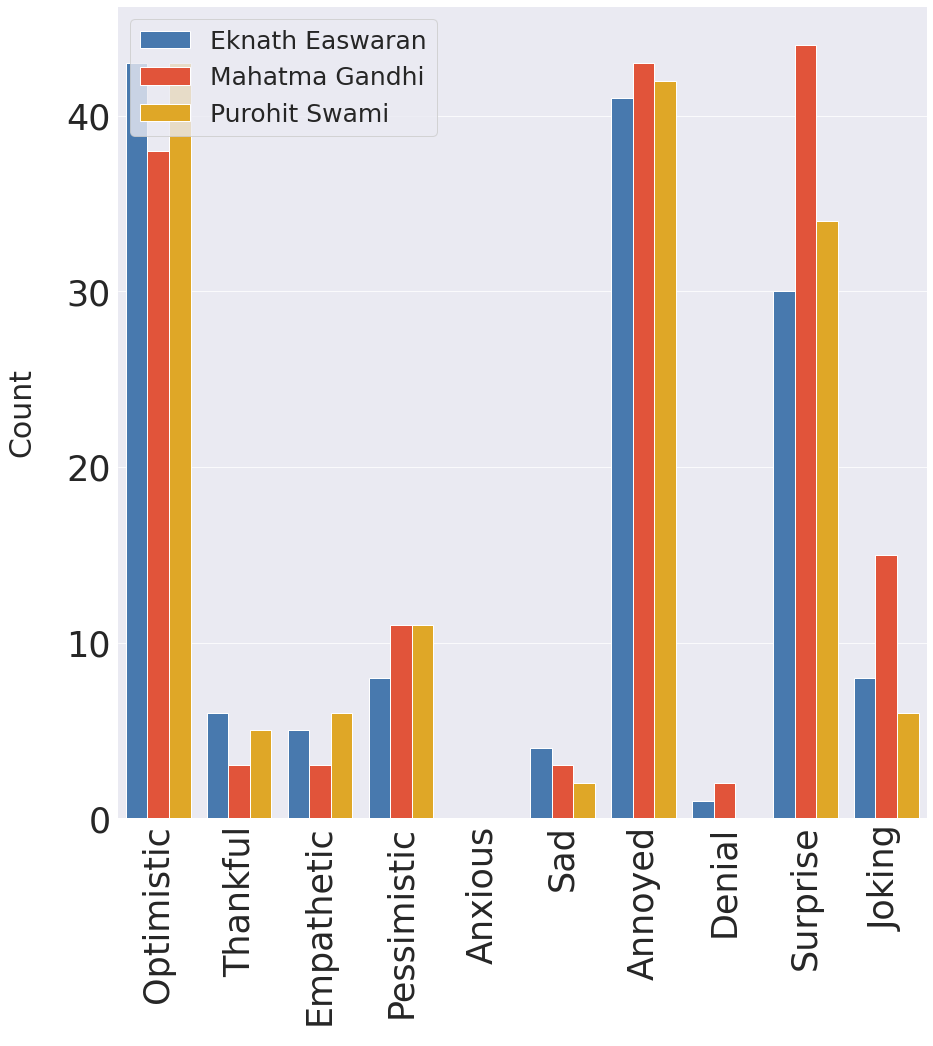}}
    
\caption{Chapter-wise sentiment analysis (Chapters 10 - 18).}
\label{fig:allchap2}
\end{figure*}

 We next present the chapter-wise sentiments for the respective translations as shown in Figures \ref{fig:chaptersent} and  \ref{fig:allchap2}.
 Figure \ref{fig:barplot_all_chapters} presents the sentiments for the entire text where we notice that sentiments such as "denial", "anxious", "thankful" and "empathetic" are the least expressed sentiments while "annoyed", "optimistic" and "surprised" are the most expressed across the three texts. In terms of chapter-wise analysis, we find that there is a major  difference in the predicted sentiments "optimistic", "pessimistic" and "surprise" in most cases shown in Figures \ref{fig:chaptersent} and  \ref{fig:allchap2}.  While Eknath Easwaran's translation leads in all positive sentiments - "optimistic", "thankful" and "empathetic", Mahatma Gandhi's translation has the highest number of "pessimistic", "annoyed", "anxious" and "surprise" sentiments. It is worthwhile to look into the text and review the exact verses and the predicted sentiments.  We select  Chapter 6 (Figure \ref{fig:chaptersent} - Panel f) and show the verses and their predicted sentiments which have a stark difference among different translations in Table \ref{tab:chapter6_verses}. The words used, grammar and language style, and the meaning associated with the verse  greatly differs across the different translators as shown in Table \ref{tab:chapter6_verses}  along with  the underlying sentiment. If we consider verse 2, we find that the given sentiments are entirely different for the different translations. This is due to the way the verse is presented to our BERT-based language model. In the case of translation by Mahatama Gandhi, the verse could be perceived as joking. Eknath Easwaran presents that same verse in a tone that is showing annoyed and surprised sentiments while Shri Purohit Swami presents it in a tone that is annoyed. We need to highlight that similar to the sentiment analysis based on tweets, the proposed model does not consider the context of the the overall chapter and hence this could be a limitation.  We provide rest of the sentiment predictions in our GitHub repository \footnote{\url{https://git.io/J9GcU}}.

\begin{figure}[htbp!]
  \centering
\subfigure[Eknath Easwaran]{ \includegraphics[width=0.99\linewidth]{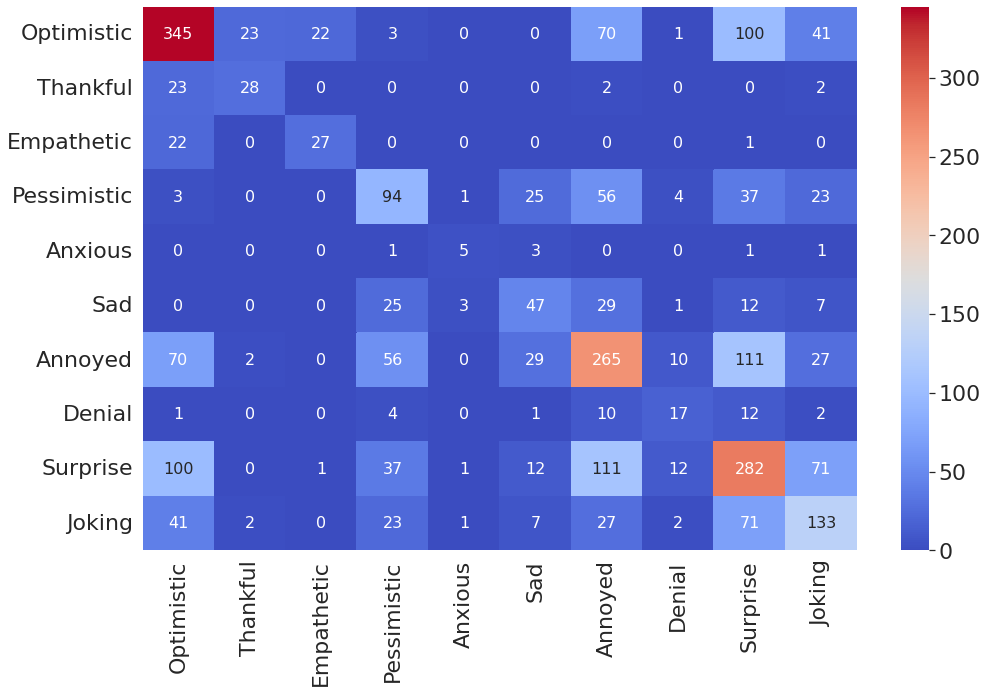}}\\
\subfigure[Mahatma Gandhi]{ \includegraphics[width=0.99\linewidth]{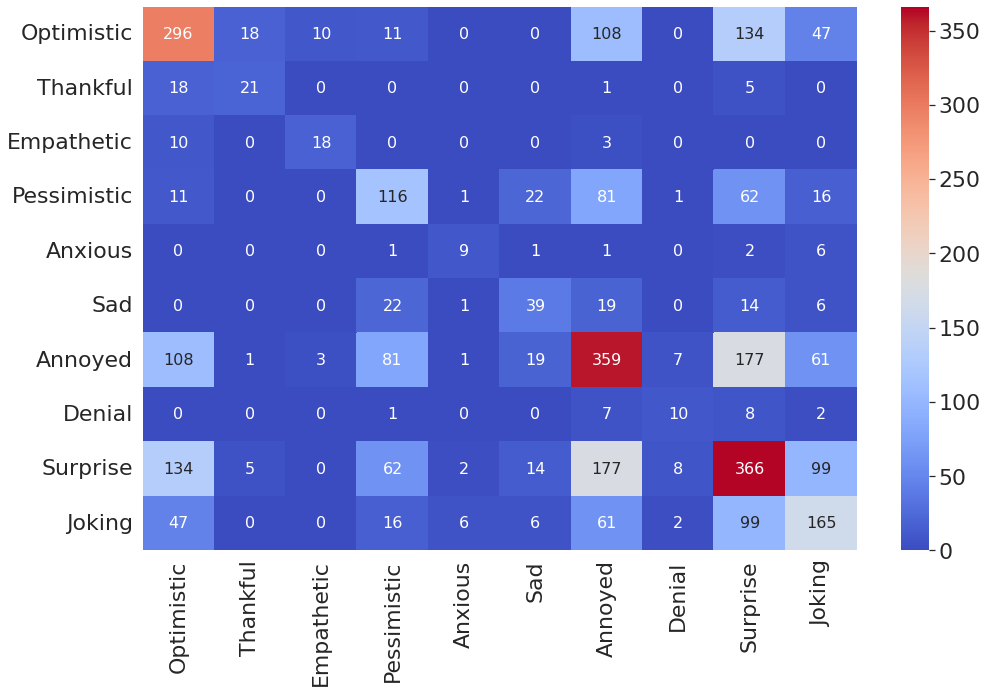}}\\

\subfigure[Shri Purohit Swami]{ \includegraphics[width=0.99\linewidth]{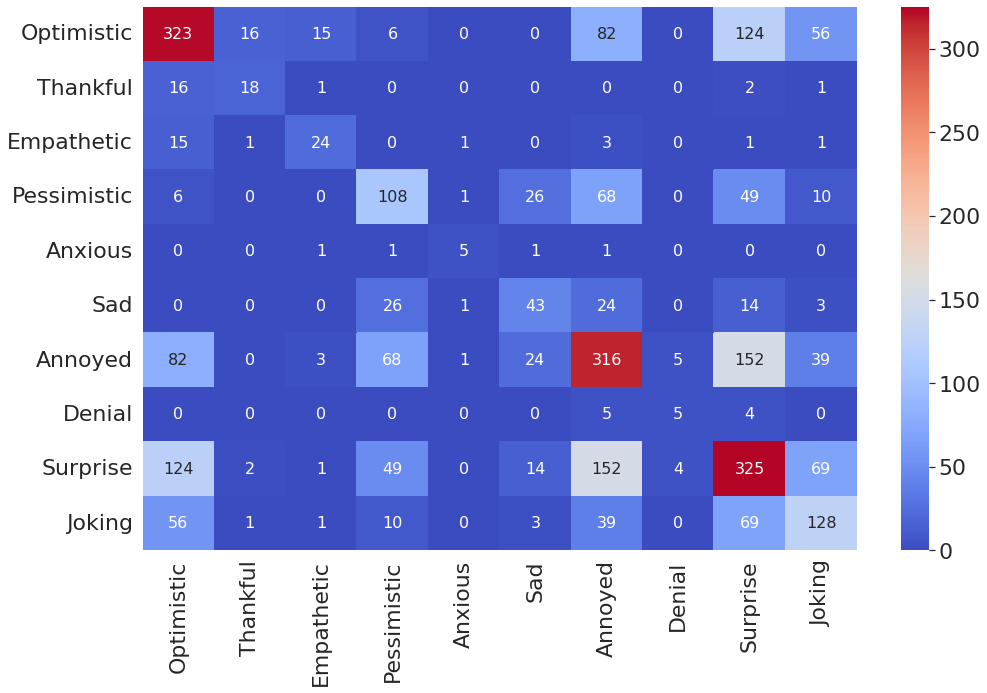}}

\caption{Heatmap showing number of occurrence of a given sentiment in relation to the rest of the sentiments of all the verses in the respective translations.}
\label{fig:heatmaps}
\end{figure}

Figure \ref{fig:heatmaps} shows the heatmap with  number of occurrence of a given sentiment in relation to the rest of the sentiments of all the verses in the respective translations. We notice that "annoyed", "optimistic" and "surprise" are the key sentiments across all the respective translations. In relation to rest of the sentiments, we find that the combination of "optimistic and annoyed", "optimistic and surprise", and "optimistic and joking" are the leading combinations.

We need to measure the similarity and diversity of the sentiments expressed with verse-by-verse comparison across the different translations. 
Table \ref{table:jaccard_score_table}  shows the Jaccard similarity score computed on the predicted sentiments for three pairs of texts for selected chapters. The score is highest for Mahatma Gandhi and Shri Purohit Swami's versions which indicates they had the highest overlap of the predicted sentiments.

\begin{table*}[htbp!]
\begin{adjustbox}{max width=\textwidth}
\begin{tabular}{|l|l|l|l|l|l|l|}
\toprule
 & Eknath Easwaran & Mahatma Gandhi & Shri Purohit Swami\\ [1ex]
\hline

Verse 2 & \begin{tabular}[c]{@{}l@{}} Therefore, Arjuna, you should understand that\\ renunciation and the performance of selfless\\ service are the same. Those who cannot \\renounce attachment to the results of their\\ work are far from the path.\end{tabular} & 

\begin{tabular}[c]{@{}l@{}} What is called sannyasa, know you to \\be yoga, O Arjuna; for none can become a\\ yogin who has not renounced selfish\\ purpose.

\end{tabular} & \begin{tabular}[c]{@{}l@{}} O Arjuna! Renunciation is in fact\\ what is called Right Action. No one\\ can become spiritual who has not renounced\\ all desire. \end{tabular} \\
\hline 

Predicted Sentiment & \begin{tabular}[c]{@{}l@{}} Annoyed, Surprise

\end{tabular} & \begin{tabular}[c]{@{}l@{}} Annoyed, Joking

\end{tabular} & \begin{tabular}[c]{@{}l@{}} Annoyed \end{tabular} \\
\hline \hline

Verse 5 & \begin{tabular}[c]{@{}l@{}} Reshape yourself through the power of \\your will; never let yourself be degraded \\by self-will. The will is the only \\friend of the Self, and the will is \\the only enemy of the Self.

\end{tabular} & \begin{tabular}[c]{@{}l@{}} By one's Self should one raise oneself, \\and not allow oneself to fall; for soul \\(Self) alone is the friend of self, \\and Self alone is self's foe.	

\end{tabular} & \begin{tabular}[c]{@{}l@{}} Let him seek liberation by the \\help of his Highest Self, and let him \\never disgrace his own Self. For that Self\\ is his only friend; yet it may also be\\ his enemy. \end{tabular} \\
\hline 

Predicted Sentiment & \begin{tabular}[c]{@{}l@{}} Optimistic, Annoyed

\end{tabular} & \begin{tabular}[c]{@{}l@{}} Pessimistic, Surprise

\end{tabular} & \begin{tabular}[c]{@{}l@{}} Optimistic, Annoyed \end{tabular} \\
\hline \hline

Verse 20 & \begin{tabular}[c]{@{}l@{}} In the still mind, in the depths of\\ meditation, the Self reveals itself. Beholding the \\Self by means of the Self, an aspirant knows \\the joy and peace of complete fulfillment.

\end{tabular} & \begin{tabular}[c]{@{}l@{}} Where thought curbed by the practice of\\ yoga completely ceases, where a man sits\\ content within himself, soul having seen soul;	

\end{tabular} & \begin{tabular}[c]{@{}l@{}} There, where the whole nature is seen in\\ the light of the Self, where the man\\ abides within his Self and is satisfied\\ there, its functions restrained by its union \\with the Divine, the mind finds rest. \end{tabular} \\

\hline

Predicted Sentiments & \begin{tabular}[c]{@{}l@{}} Optimistic

\end{tabular} & \begin{tabular}[c]{@{}l@{}} Pessimistic, Sad, Surprise

\end{tabular} & \begin{tabular}[c]{@{}l@{}} Optimistic, Surprise \end{tabular} \\

\hline\hline

Verse 32 & \begin{tabular}[c]{@{}l@{}} ARJUNA: O Krishna, the stillness of divine union\\ which you describe is beyond my comprehension.\\ How can the mind, which is so restless,\\ attain lasting peace?

\end{tabular} & \begin{tabular}[c]{@{}l@{}} 	I do not see, O Krishna, how this\\ yoga, based on the equal-mindedness that\\ You have expounded to me, can \\steadily endure, because of fickleness\\ (of the mind).	

\end{tabular} & \begin{tabular}[c]{@{}l@{}} Arjuna said: I do not see how I can \\attain this state of equanimity which You\\ has revealed, owing to the restlessness\\ of my mind.	 \end{tabular} \\

\hline

Predicted Sentiments & \begin{tabular}[c]{@{}l@{}} Surprise

\end{tabular} & \begin{tabular}[c]{@{}l@{}} Pessimistic, Annoyed

\end{tabular} & \begin{tabular}[c]{@{}l@{}} {Surprise} \end{tabular} \\

\hline
\hline

\end{tabular}
\end{adjustbox}
 \caption{Some of the least similar verses in Chapter 6 according to predicted sentiments.}
    \label{tab:chapter6_verses}
\end{table*}

\begin{figure*}[hbbp!]
  \centering
  \includegraphics[width=0.8\textwidth]{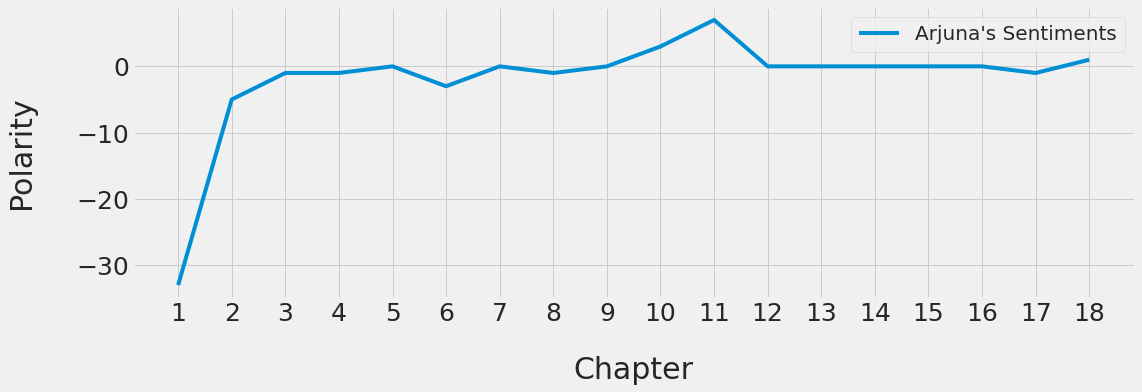}
  \caption{Arjuna's sentiments over time (chapters) in the translation by Eknath Easwaran.}
  \label{arjuna_sentiments}
  
\end{figure*}

\begin{figure*}[hbbp!]
  \centering
  \includegraphics[width=0.7\textwidth]{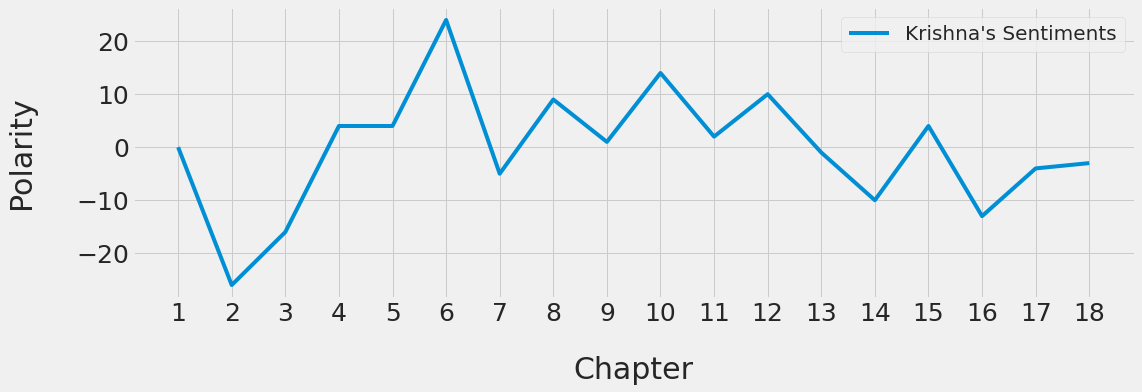}
  \caption{Lord  Krishna's sentiments over time (chapters) in the translation by Eknath Easwaran.}
  \label{krishna_sentiments}
  
\end{figure*}

\begin{table}[!htb]
\centering 
\begin{tabular}{| c | c | c | c |} 
\hline\hline 
 & Easwaran-Gandhi & Purohit-Easwaran & Gandhi-Purohit \\ [1ex] 
\hline 
Chapter 3 & 0.604 & 0.506 & 0.684 \\ [0.5ex] 
Chapter 5 & 0.568 & 0.520 & 0.614 \\ [0.5ex]
Chapter 7 & 0.559 & 0.486 & 0.574 \\ [0.5ex]
Chapter 8 & 0.547 & 0.503 & 0.568 \\ [0.5ex]
Chapter 9 & 0.501 & 0.486 & 0.565  \\ [0.5ex]
Chapter 10 & 0.523 & 0.507 & 0.562
 \\ [0.5ex]
Chapter 11 & 0.507 & 0.488 & 0.527 \\ [0.5ex]
Chapter 12 & 0.500 & 0.483 & 0.518 \\ [0.5ex]
Chapter 15 & 0.494 & 0.490 & 0.516 \\ [0.5ex]
Chapter 16 & 0.500 & 0.495 & 0.527 \\ [0.5ex] 
Chapter 17 & 0.510 & 0.501 & 0.536 \\ [1ex]
\hline 
* & 0.526 & 0.497 &  0.563 \\ [1 ex]

\hline
\end{tabular}
\caption{Sentiment analysis of selected pairs of translations  with Jaccard similarity score for predicted sentiments for select chapters.  We provide the mean  of the scores at the bottom (*).}
\label{table:jaccard_score_table}  
\end{table}

\subsection{Semantic Analysis}

Next, we provide analysis of the translation    using semantic analysis of verse-by-verse comparison of the text. We encode the verses   using the MPNet-base model and report the  cosine similarity score  for every chapter.
Table  \ref{table:cosine_similarity_score_table} shows the verse by verse   cosine similarity score for the translations grouped by the chapter. We provide the mean and standard deviation of the score for the respective verses in each chapter.  We observe that Chapter 17 is semantically least similar and Chapter 10 is most similar. Further, in the pair-wise comparison, Mahatma Gandhi and Shri Purohit Swami's translation are most similar to each other. It is important to note that these two translations also have the highest Jaccard similarity score for the predicted sentiments. We present some of the semantically least similar verses in Table \ref{tab:semantic_least_similar_verses}. In  Chapter 5 -- Verse 5, Eknath Easwaran interprets the goals of knowledge and service to be the same, Mahatma Gandhi sees schools of  Sankhya and Yoga philosophy as one, and Shri Purohit Swami interprets wisdom (Jnana) and right action (Dharma) as one. In this case, Mahatma Gandhi's verse is least similar to the other two (Score 1 and Score 2 are both lower than Score 3), primarily because it uses the Sanskrit terms of Sankhya and Yoga. Whenever, a particular translation uses a Sanskrit term, it obtains a lower cosine similarity score with respect to the other two as BERT-based models have been primarily trained on a large corpus of English texts. In Chapter 16 -- Verse 4, we observe that there is a difference in the devilish qualities being discussed. Some of the qualities unique to each text are: cruelty (Eknath Easwaran), pretentiousness, coarseness (Mahatma Gandhi) and pride, insolence (Shri Purohit Swami). The difference in choice of words may be due to the different eras in which these translations were created, leading to the use of  words prevalent at that point of time. Next, we present semantically most similar verses in Table \ref{tab:semantic_most_similar_verses}. In  Chapter 3 -- Verse 4, we observe that although the choice of words are different, the meaning conveyed is similar. We find that the BERT-based language model assigns a high similarity score to all the three pairs. In  Chapter 9 -- Verse 18, the choice of words used by Mahatma Gandhi and Shri Purohit Swami is almost identical and hence, we obtain a high similarity score (Score 2).

Furthermore, we review the semantic score by presenting the actual verses from the translated versions of a selected chapter. We choose Chapter 12 since this chapter has one of the least verses, it is easier for us to include it in the paper. We publish the scores for the rest of the chapters  in our GitHub repository \footnote{\url{https://git.io/JSyPB}}. In Table \ref{tab:semantic_similarity_chapter_12}, we present all the verses taken from Chapter 12 with the cosine similarity score. We also present the mean and the standard deviation of the   scores to provide an indication about the overall semantic similarity of the verses for the chapter for the comparison of selected translations. 

\begin{figure*}[htbp!]
    \centering
    
\subfigure[Chapter 1]{\label{fig:chapter1_all_text_umap}\includegraphics[width=.45\linewidth]{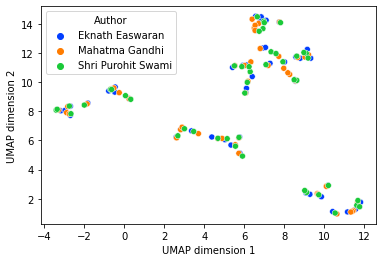}}
\subfigure[Chapter 4 ]{\label{fig:chapter4_all_text_umap}\includegraphics[width=.45\linewidth]{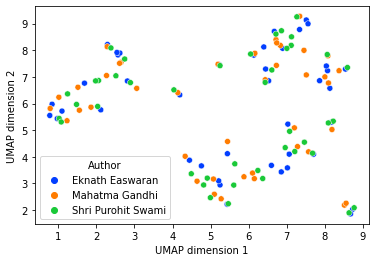}}
\subfigure[Chapter 7 ]{\label{fig:chapter7_all_text_umap}\includegraphics[width=.45\linewidth]{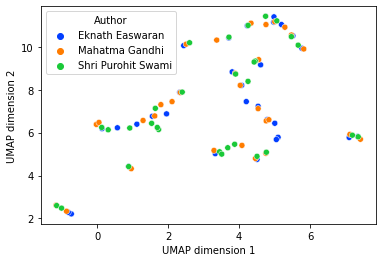}}
\subfigure[Chapter 10 ]{\label{fig:chapter10_all_text_umap}\includegraphics[width=.45\linewidth]{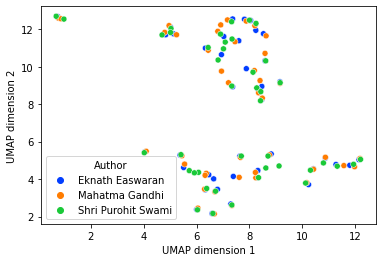}}
\subfigure[Chapter 13 ]{\label{fig:chapter13_all_text_umap}\includegraphics[width=.45\linewidth]{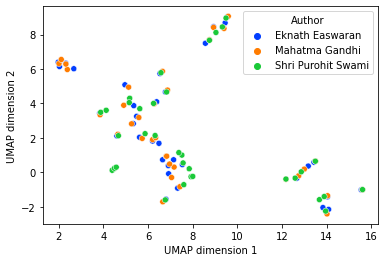}}
\subfigure[Chapter 17 ]{\label{fig:chapter17_all_text_umap}\includegraphics[width=.45\linewidth]{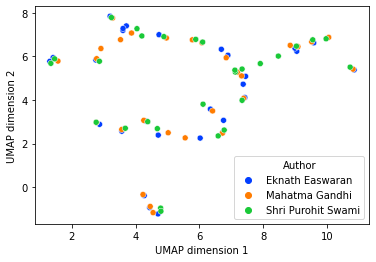}}
    
\caption{Chapter-wise UMAP embedding visualization.}
\label{fig:chapter_wise_UMAP}
\end{figure*}

 We visualise the encoded verses in two dimensions using UMAP  in Figure \ref{fig:chapter_wise_UMAP}. We present the visualizations for one in every three chapters. We do not intend to find separate clusters for each translation as it is evident from Table \ref{table:cosine_similarity_score_table} that the verses from each translation are semantically very similar to each other. However, we do find some interesting clusters which represent a sequence of events or consecutive verses in the text. For example, in Figure \ref{fig:chapter1_all_text_umap}, the distinct clusters represent separate events. In Chapter 1, these include verses describing Sanjaya's description of battlefield to Dhritarashtra, Arjuna moving his chariot to the center of the battlefield, Arjuna expressing his grief to Krishna. In Chapter 4 (Figure \ref{fig:chapter4_all_text_umap}), we observe 3 clusters. The left-most cluster includes verses describing different forms of sacrifice, the bottom-most cluster comprises verses where Lord Krishna emphasises renunciation of oneself from fruits of action, and the third cluster includes the conversation between Krishna and Arjuna from the first ten verses of the chapter. In Chapter 10 (Figure \ref{fig:chapter10_all_text_umap}), the lower cluster comprises the verses in which Lord Krishna describes his divine attributes to Arjuna, and the upper cluster features the rest of the conversation between Krishna and Arjuna in the chapter (primarily first 20 verses).

Furthermore, the keywords in the text/chapter play a vital role in semantic similarity.  It is important to note that most keywords aptly signify the themes of the chapters. This helps us to quickly assess how different authors differ in their interpretation of key concepts. Table \ref{tab:keywords_all_chapters} shows the keywords obtained for all the  chapters for the respective translations.  In order to get a better perspective, we need to review the keywords from a selected chapter. 
 Table \ref{table:keywords_chapter_12} shows the keywords from Chapter 12 and their associated scores obtained using KeyBert.

 Table \ref{table:keywords_chapter_12}   shows the keywords extracted from Chapter 12. This chapter starts with Arjuna asking Lord Krishna who are the better practitioners of yoga: those who love him dearly or those who seek the eternal reality.   Lord Krishna responds that worshipping the eternal reality (Brahman) \cite{junghare2011unified,chaudhuri1954concept} is extremely difficult for the humans. So, he suggests Arjuna to meditate and be devoted to him while engaging in selfless action (karma yoga). We note again that Lord Krishna is the  avatar of Lord Vishnu (supreme consciousness) who is equivalent to the Abrahamic notion of God. A steady and  stable  state of mind can only be achieved by humans with the practice of meditation. It is pertinent to note the KeyBERT has extracted "meditate" as one of the top 10 keywords in Shri Purohit Swami's translation. Lord Krishna further says that the devotees who focus their mind with unflinching devotion are very dear to him (Chapter 12 -- Verse 20). 'Devotion' and 'devotee' marks another central theme (topic) that has been correctly extracted, which also is the top keyword in all three translations. Furthermore,  words such as "eternal", "unchanging", "omnipresent" are used to refer to the  Atman (self/qualia). Arjuna and Krishna, though not themes, appear extensively in the chapter and therefore, they have been also  extracted.  We report  the cumulative score where we have added the cosine similarity scores of the keywords in batches of 15 verses each (Table \ref{table:keywords_chapter_12}).

\begin{figure}[htbp!]
  \centering
\subfigure[Eknath Easwaran]{\label{fig:UMAP_eknath_easwaran}\includegraphics[width=.850\linewidth]{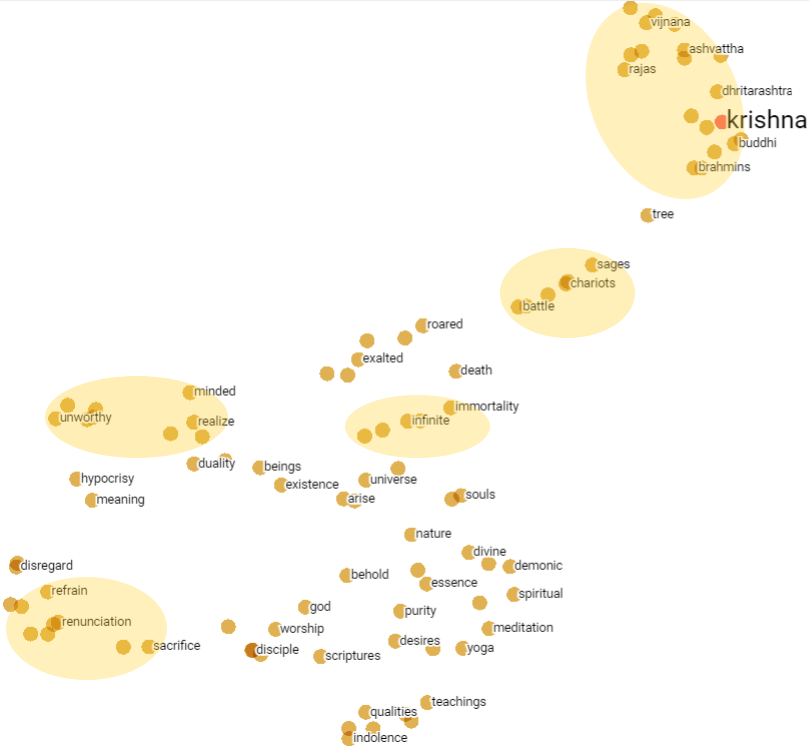}}\\
\subfigure[Mahatma Gandhi]{\label{fig:UMAP_mahatma_gandhi}\includegraphics[width=.850\linewidth]{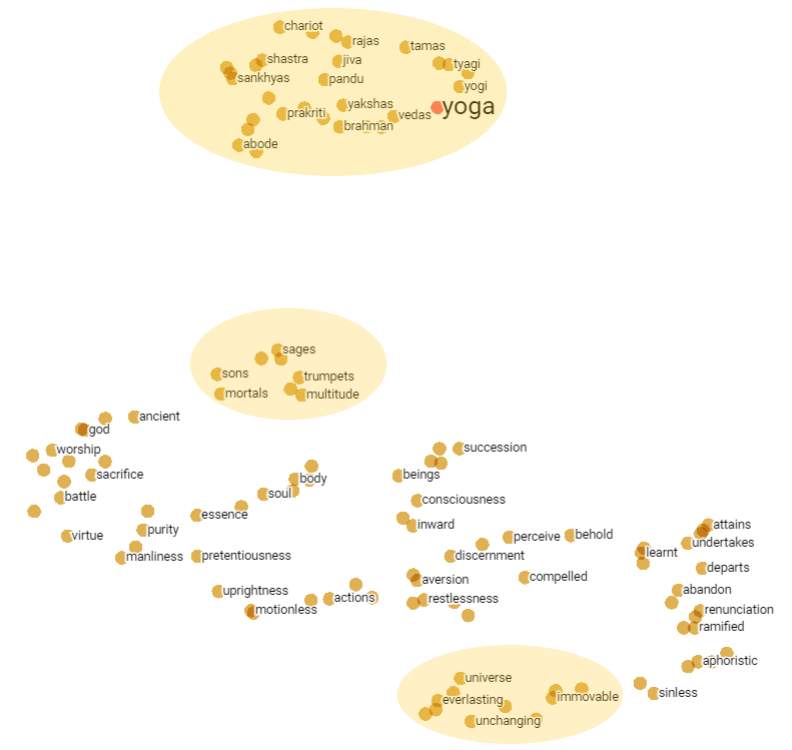}}

\subfigure[Shri Purohit Swami]{\label{fig:UMAP_shri_purohit_swami}\includegraphics[width=.850\linewidth]{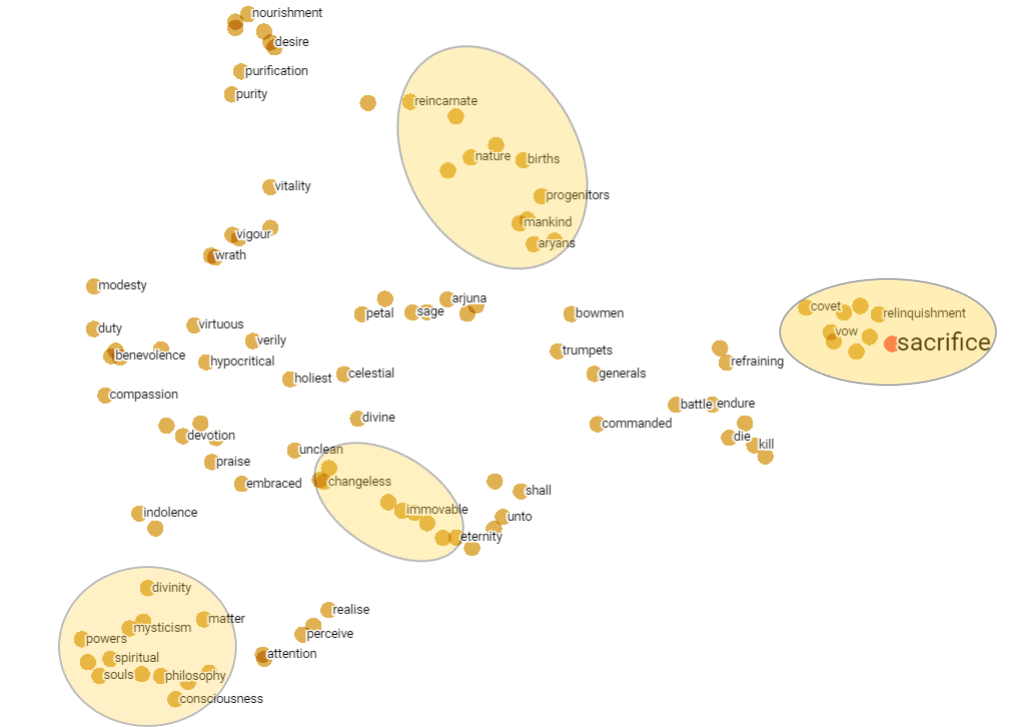}}

\caption{Visualisation of UMAP embedding of keywords from the respective translations. }
\label{UMAP_EMBEDDING}
\end{figure}

In order to provide further visualisation of the keywords and how they are related to each other, we use dimension reduction and plot the first two components of the latent vectors. We use uniform manifold approximation and projection (UMAP) \cite{mcinnes2018umap} for dimension reduction.
 We obtained the UMAP embeddings  by encoding top 10 keywords from all chapters using the MPNet sentence embedding model. The keywords which appeared in more than one chapter were removed. In these visualisations, we are primarily interested in analyzing clusters of data as UMAP algorithm is good at preserving the local structure by grouping neighboring data points together.  Figure \ref{UMAP_EMBEDDING} presents the visualisation (scatter plot) of UMAP embedding.  The cluster of keywords that provide important semantic information, with respect to their use in the Bhagavad Gita, are shaded in yellow. In Eknath Easwaran's translation, (Figure \ref{fig:UMAP_eknath_easwaran}) the words associated  to the impending battle such  "archers", "chariots", "war", "armies", "battle" form a cluster.  Also, "roared", "death" and "killed" are close to this cluster in the embedding space. The keywords in the philosophy of giving away (the central theme in Chapter 5 -- Renounce and Rejoice) also form a cluster. These include "sacrifice" (and "sacrifices"), "renounce", "renouncing", "renunciation", "abstaining", "surrendering", all of which have a similar underlying meaning. One of the central themes in Easwaran's translation - \textit{selfless service} is captured by words such as  "self-realization", "selflessly"  ("selfless"), "realize" which cluster together. However, "selfish", "selfishly" also are a part of this cluster though they convey the opposite meaning. The terms "eternal", "immutable", "changeless" and "immortality"  which are closely associated with   qualia (Atman) appear as a cluster.  The keywords representing the protagonists and associates in the Bhagavad Gita ("Krishna", "Arjuna", "Duryodhana", "Dhritarashtra") and their qualities such as  "sattva", "dharma", "tyaga", "rajas" and "tamas" form a cluster in the upper portion of the embedding space.  
 
 In Mahatma Gandhi's translation (Figure \ref{fig:UMAP_mahatma_gandhi}),  "rajas, tamas" (both undesirable qualities) are very close to each other. Furthermore, most of the prominent personalities representing the righteous side of the war - "Krishna", "Arjuna", "Bhima" (Arjuna's brother), "Pandu" (Arjuna's father), "Subhadra" (Arjuna's wife) and  some of their qualities and attributes such as "tyagi", "yogi", "sattva", "rajas", "tamas" form a cluster. Moreover, 
"Ashvattha" \cite{haberman2013people} which is a sacred fig tree for Hindus is very close to the keyword "tree".
The words related to   the Atman (qualia) - "omnipresent", "transcending", "unchanging", "imperishable", "everlasting" "immovable" form a cluster near the bottom of the plot. The entities involved in or pertaining to the impending war of Kurukshetra - "bowmen", "vestments", "trumpets", "multitude" (referring to huge number of armed men on both sides), "mortals", "sons" are also found to cluster together. Finally, "lord", "devotion", "devotee", "worship", "reverence" also form a cluster.   In Shri Purohit Swami's translation, Figure \ref{fig:UMAP_shri_purohit_swami}, the attributes of soul (as described in the Bhagavad Gita) form a cluster. These attributes that include - "divinity", "mysticism", "spirit", "power", "consciousness", "philosophy", "wisdom", "knowledge" form a cluster with "souls". Sacrifice is another key theme in Shri Purohit Swami's translation. In our visualization, "sacrifice", "relinquishment", "renunciation", and "vow" cluster together. The idea of the eternity of the soul is captured by the words - "omniscient", "omnipresent", "immovable", "changeless", "eternal". While the eternity of soul is a widely discussed theme in the Bhagavad Gita, so is the idea of rebirth. "Progenitors", "reborn", "reincarnation", "births", "creation", "nature", "mankind" are the keywords closely related to this idea that cluster together. We have not shown all the keywords due to limited space.  It is interesting to note that in the respective UMAP embeddings, the points representing Arjuna and Lord Krishna are always closest to each other. Please note that the bigger sizes of the words "krishna" , "yoga" and "sacrifice"    do not imply any semantic importance (Figure \ref{fig:UMAP_eknath_easwaran}, Figure \ref{fig:UMAP_mahatma_gandhi}, and Figure \ref{fig:UMAP_shri_purohit_swami}).

\begin{table*}[!htb]
\centering 
\begin{tabular}{| c | c | c | c |} 
\hline\hline 
 & Easwaran vs Gandhi & Purohit vs Easwaran & Gandhi vs Purohit \\ [1ex] 
\hline 
Chapter 3 & 0.627    (0.133) & 0.608 (0.146) & 0.694 (0.114) \\ [0.5ex] 
Chapter 5 & 0.629 (0.129) & 0.637 (0.105) & 0.628 (0.122)\\ [0.5ex]
Chapter 7 & 0.698 (0.144) & 0.668 (0.147) & 0.691 (0.148) \\ [0.5ex]
Chapter 8 & 0.662 (0.123) & 0.650 (0.081) & 0.673 (0.124) \\ [0.5ex]
Chapter 9 & 0.681 (0.126) & 0.686 (0.109) & 0.731 (0.094)\\ [0.5ex]
Chapter 10 & 0.759 (0.096) & 0.747 (0.113) & 0.733 (0.123)
 \\ [0.5ex]
Chapter 11 & 0.705 (0.109) & 0.731 (0.101) & 0.721 (0.110)\\ [0.5ex]
Chapter 12 & 0.609 (0.120) & 0.633 (0.097) & 0.724 (0.096) \\ [0.5ex]
Chapter 15 & 0.686 (0.116) & 0.675 (0.110) & 0.657 (0.067)\\ [0.5ex]
Chapter 16 & 0.662 (0.096) & 0.667 (0.100) & 0.691 (0.116)\\ [0.5ex] 
Chapter 17 & 0.654 (0.111) & 0.591 (0.132) & 0.642 (0.132) \\ [1ex]
\hline 
* & 0.670 (0.119) & 0.663 (0.113)& 0.689 (0.113) \\ [1 ex]

\hline
\end{tabular}
\caption{Semantic analysis using cosine similarity for selected chapters from selected pairs of the translations. Note that the mean is given with standard deviation (in brackets) for all the verses in the respective chapters at the bottom (*). }
\label{table:cosine_similarity_score_table} 
\end{table*}

\begin{table*}[htbp!]
\resizebox{\textwidth}{!}{%
\begin{tabular}{|l|l|l|l|l|l|l|l|}
\toprule
Chapter & Verse & Eknath Easwaran & Mahatma Gandhi & Shri Purohit Swami & Score 1 & Score 2 & Score 3\\ [1ex]
\hline
5 & 5 & \begin{tabular}[c]{@{}l@{}} The goal of knowledge and\\ the goal of service are the same;\\ those who fail to see this\\ are blind.

\end{tabular} & \begin{tabular}[c]{@{}l@{}}The goal that the sankhyas \\attain is also reached by the yogins.\\ He sees truly  who sees both sankhya\\ and yoga as one.

\end{tabular} & \begin{tabular}[c]{@{}l@{}} The level which is reached by\\ wisdom is attained through right \\action as well. He who perceives \\that the two are one,\\ knows the truth.  \end{tabular}  & 0.295 & 0.338 & 0.517\\
\hline 

9 & 22 & \begin{tabular}[c]{@{}l@{}} Those who worship me and \\meditate on me constantly, without\\ any other thought  I will\\ provide for all their needs. 

\end{tabular} & \begin{tabular}[c]{@{}l@{}}  As for those who worship\\ Me, thinking on Me alone \\and nothing else, ever  attached\\ to Me, I bear the burden\\ of getting them what they need.   

\end{tabular} & \begin{tabular}[c]{@{}l@{}}  But if a man will meditate on Me\\ and Me alone, and will worship\\ Me always and everywhere,\\ I will take upon Myself\\ the fulfillment of his aspiration,\\ and I will safeguard whatsoever\\ he shall attain.  \end{tabular} & 0.571 & 0.527 & 0.528 \\
\hline

15 & 11 & \begin{tabular}[c]{@{}l@{}}     Those who strive resolutely on\\ the path of yoga see the\\ Self within. The thoughtless,\\ who strive imperfectly, do not. 

\end{tabular} & \begin{tabular}[c]{@{}l@{}}  Yogins who strive see Him seated\\ in themselves; the witless ones who have\\  not cleansed themselves to see Him\\ not, even though they strive.   

\end{tabular} & \begin{tabular}[c]{@{}l@{}}   The saints with great effort find\\ Him within themselves; but not the\\ unintelligent, who in spite of every\\ effort cannot control their minds. 
\end{tabular} & 0.450 & 0.629 & 0.367 \\
\hline

16 & 4 & \begin{tabular}[c]{@{}l@{}}    Other qualities, Arjuna, make a person\\ more and more inhuman: hypocrisy, arrogance,\\ conceit, anger, cruelty, ignorance. 

\end{tabular} & \begin{tabular}[c]{@{}l@{}}   Pretentiousness, arrogance, self-conceit, wrath, \\coarseness, ignorance—these  are to be found\\ in one born with the devilish\\ heritage.   

\end{tabular} & \begin{tabular}[c]{@{}l@{}}    Hypocrisy, pride, insolence, cruelty, \\ignorance belong to him who is\\ born of the godless qualities. 
\end{tabular} & 0.447 & 0.586 & 0.520 \\
\hline

17 & 18 & \begin{tabular}[c]{@{}l@{}}   Disciplines practiced in order to \\gain respect, honor, or admiration\\ are rajasic; they are undependable and\\ transitory in their effects. 

\end{tabular} & \begin{tabular}[c]{@{}l@{}}  Austerity which is practiced \\with an eye to gain praise, honour\\ and homage  and for ostentation is said\\ to be rajasa; it is fleeting \\and unstable.   

\end{tabular} & \begin{tabular}[c]{@{}l@{}}   Austerity coupled with hypocrisy or\\ performed for the sake of \\self-glorification, popularity or vanity, comes\\ from Passion, and its result is \\always doubtful and temporary.  \end{tabular} & 0.440 & 0.740 & 0.212 \\
\hline

\hline

\end{tabular}
}
 \caption{Semantically least similar verses.  The cosine similarity (score) is given with the following combinations. Note that the comparisons are defined by Eknath Easwaran vs Mahatma Gandhi (Score 1),   Mahatma Gandhi vs Shri Purohit Swami (Score 2), and   Shri Purohit Swami vs Eknath Easwaran (Score 3).}
    \label{tab:semantic_least_similar_verses}
\end{table*}

\begin{table*}[htbp!]
\resizebox{\textwidth}{!}{%
\begin{tabular}{|l|l|l|l|l|l|l|l|}
\toprule
Chapter & Verse & Eknath Easwaran & Mahatma Gandhi & Shri Purohit Swami & Score 1 & Score 2 & Score 3\\ [1ex]
\hline

3 & 4 & \begin{tabular}[c]{@{}l@{}} One who shirks action does not\\ attain freedom; no one can gain\\ perfection by abstaining from work.

\end{tabular} & \begin{tabular}[c]{@{}l@{}}  Never does man enjoy freedom\\ from action by not undertaking \\action, nor does he attain that \\freedom by mere renunciation of action. 

\end{tabular} & \begin{tabular}[c]{@{}l@{}} No man can attain freedom\\ from activity by refraining from \\action; nor can he reach perfection\\ by merely refusing to act. \end{tabular} & 0.753 & 0.879 & 0.837 \\
\hline

7 & 9 & \begin{tabular}[c]{@{}l@{}} I am the sweet fragrance in\\ the earth and the radiance of\\ fire; I am the life in\\ every creature and the striving\\ of the spiritual aspirant.

\end{tabular} & \begin{tabular}[c]{@{}l@{}}  I am the sweet fragrance\\ in earth; the brilliance in\\ fire; the life in all beings;\\ and the austerity in ascetics.   

\end{tabular} & \begin{tabular}[c]{@{}l@{}}  I am the Fragrance of earth,\\ the Brilliance of fire. I am \\the Life Force in all beings,\\ and I am the Austerity of the ascetics. \end{tabular} & 0.909 & 0.917 & 0.856 \\
\hline

9 & 18 & \begin{tabular}[c]{@{}l@{}} I am the goal of life,\\ the Lord and support of all, \\the inner witness, the abode of\\ all. I am the only refuge,\\ the one true friend; I am the\\ beginning, the staying, and the\\ end of creation; I am the \\womb and the eternal seed.

\end{tabular} & \begin{tabular}[c]{@{}l@{}}I am the Goal, the Sustainer,\\ the Lord, the Witness, the Abode,\\ the Refuge, the Friend; the Origin,\\ the End the Preservation, the Treasure\\ house, the Imperishable Seed.

\end{tabular} & \begin{tabular}[c]{@{}l@{}} I am the Goal, the Sustainer,\\ the Lord, the Witness, the Home,\\ the Shelter, the Lover and\\ the Origin; I am Life and Death;\\ I am the Fountain and the Seed Imperishable.	  \end{tabular}  & 0.839 & 0.907 & 0.871\\
\hline 

12 & 18 & \begin{tabular}[c]{@{}l@{}}   That devotee who looks upon\\ friend and foe with equal regard,\\ who is not buoyed up by praise \\nor cast down by blame, alike\\ in heat and cold, pleasure and\\ pain, free from selfish attachments,

\end{tabular} & \begin{tabular}[c]{@{}l@{}}Who is same to foe and friend,\\ who regards alike respect and disrespect,\\ cold and heat, pleasure and pain,\\ who is free from attachment;

\end{tabular} & \begin{tabular}[c]{@{}l@{}}  He to whom friend and foe\\ are alike, who welcomes equally honour\\ and dishonour, heat and cold, pleasure\\ and pain, who is enamoured of nothing, \end{tabular} & 0.827 & 0.885 & 0.820 \\
\hline

\hline

\end{tabular}
}
 \caption{Semantically most similar verses.  The cosine similarity (score) is given with the following combinations.  Note that the comparisons are defined by Eknath Easwaran vs Mahatma Gandhi (Score 1),   Mahatma Gandhi vs Shri Purohit Swami (Score 2), and   Shri Purohit Swami vs Eknath Easwaran (Score 3).}
    \label{tab:semantic_most_similar_verses}
\end{table*}

\begin{table*}[htbp!]
\resizebox{\textwidth}{!}{%
\begin{tabular}{|l|l|l|l|l|l|l|}
\toprule
Verse & Eknath Easwaran & Mahatma Gandhi & Shri Purohit Swami & Score 1 & Score 2 & Score 3\\ [1ex]
\hline
1 & \begin{tabular}[c]{@{}l@{}} ARJUNA: Of those steadfast \\devotees who love you and \\those who seek you as the \\eternal formless Reality, who \\are the more established in yoga?

\end{tabular} & \begin{tabular}[c]{@{}l@{}}Of the devotees who thus worship\\ You, incessantly attached, and \\those who  worship the \\Imperishable Unmanifest, which are \\the better yogins?  The Lord Said:

\end{tabular} & \begin{tabular}[c]{@{}l@{}}“Arjuna asked: My Lord! Which are the better\\ devotees who worship You, those who try \\to know You as a Personal God, or those\\ who worship You as Impersonal and Indestructible? \end{tabular}  & 0.414 & 0.589 & 0.573\\
\hline 

2 & \begin{tabular}[c]{@{}l@{}}   
KRISHNA:  Those who set their hearts\\ on me and worship me with unfailing\\ devotion and faith are more \\established in yoga.

\end{tabular} & \begin{tabular}[c]{@{}l@{}}  Those I regard as the best yogins\\ who, riveting their minds on Me, ever\\  attached, worship Me, with the\\ highest faith.      

\end{tabular} & \begin{tabular}[c]{@{}l@{}}  Lord Shri Krishna replied: Those\\ who keep their minds fixed on\\ Me, who worship Me always with \\unwavering faith and concentration; these\\ are the very best.   \end{tabular}  & 0.490 & 0.575 & 0.663\\
\hline 

3 & \begin{tabular}[c]{@{}l@{}}   
As for those who seek the\\ transcendental Reality, without name,\\ without form, contemplating the Unmanifested,\\ beyond the reach of thought and \\of feeling, 

\end{tabular} & \begin{tabular}[c]{@{}l@{}}But those who worship the\\ Imperishable, the indefinable, the \\Unmanifest, the  Omnipresent, the \\Unthinkable, the Rock-seated, the Immovable,\\ the  Unchanging,   

\end{tabular} & \begin{tabular}[c]{@{}l@{}} Those who worship Me as the\\ Indestructible, the Undefinable,\\ the Omnipresent, the Unthinkable,\\ the Primeval, the Immutable and\\ the Eternal;  \end{tabular}  & 0.579 & 0.826 & 0.488\\
\hline

4 & \begin{tabular}[c]{@{}l@{}}   
 with their senses subdued and\\ mind serene and striving for\\ the good of all beings, they\\ too will verily come unto me. 

\end{tabular} & \begin{tabular}[c]{@{}l@{}}  Keeping the whole host of senses\\ in complete control, looking on all\\ with an  impartial eye, engrossed in\\ the welfare of all beings—these come\\ indeed to Me.      

\end{tabular} & \begin{tabular}[c]{@{}l@{}}  Subduing their senses, viewing all \\conditions of life with the same eye,\\ and working for the welfare of all\\ beings, assuredly they come to Me.  \end{tabular}  & 0.533 & 0.619 & 0.742\\
\hline

5 & \begin{tabular}[c]{@{}l@{}}   
   Yet hazardous and slow is the\\ path to the Unrevealed, difficult\\ for physical creatures to tread. 

\end{tabular} & \begin{tabular}[c]{@{}l@{}}  Greater is the travail of those\\ whose mind is fixed on the\\ Unmanifest; for it is  hard for \\embodied mortals to gain the Unmanifest—Goal.

\end{tabular} & \begin{tabular}[c]{@{}l@{}}  But they who thus fix their\\ attention on the Absolute and Impersonal\\ encounter greater hardships, for it is\\ difficult for those who possess a body\\ to realise Me as without one.   \end{tabular}  & 0.430 & 0.622 & 0.426\\
\hline

6 & \begin{tabular}[c]{@{}l@{}}   
    But they for whom I am the\\ supreme goal, who do all work \\renouncing self for me and meditate on\\ me with single-hearted devotion, 

\end{tabular} & \begin{tabular}[c]{@{}l@{}}    But those who casting all their\\ actions on Me, making Me their all \\in all,  worship Me with the meditation\\ of undivided devotion,   

\end{tabular} & \begin{tabular}[c]{@{}l@{}}   Verily, those who surrender their actions\\ to Me, who muse on Me, worship Me\\ and meditate on Me alone, with no\\ thought save of Me,    \end{tabular}  & 0.776 & 0.799 & 0.725\\
\hline

7 & \begin{tabular}[c]{@{}l@{}}   
     these I will swiftly rescue from\\ the fragments cycle of birth and death,\\ for their consciousness has entered into me. 

\end{tabular} & \begin{tabular}[c]{@{}l@{}}      Of such, whose thoughts are centered \\on Me, O Arjuna, I become before \\long the  Deliverer from the ocean \\of this world of death.       

\end{tabular} & \begin{tabular}[c]{@{}l@{}}    O Arjuna! I rescue them from the ocean\\ of life and death, for their minds\\ are fixed on Me.     \end{tabular}  & 0.546 & 0.752 & 0.560\\
\hline

8 & \begin{tabular}[c]{@{}l@{}}   
       Still your mind in me, still your\\ intellect in me, and without doubt\\ you will be united with me forever.  

\end{tabular} & \begin{tabular}[c]{@{}l@{}}        On Me set your mind, on Me rest\\ your conviction; thus without doubt shall\\ you  remain only in Me hereafter.          

\end{tabular} & \begin{tabular}[c]{@{}l@{}}     Then let your mind cling only\\ to Me, let your intellect abide in\\ Me; and without doubt you shall live\\ hereafter in Me alone.      \end{tabular}  & 0.632 & 0.788 & 0.669\\
\hline

9 & \begin{tabular}[c]{@{}l@{}}   
        If you cannot still your mind\\ in me, learn to do so through the\\ regular practice of meditation. 
        
\end{tabular} & \begin{tabular}[c]{@{}l@{}}   If you can not set your mind steadily\\ on Me, then by the method of constant  practice\\ seek to win Me, O Arjuna.   

\end{tabular} & \begin{tabular}[c]{@{}l@{}}   But if you can not fix your mind \\firmly on Me, then, My beloved friend,\\ try to do so by constant practice.    \end{tabular}  & 0.583 & 0.622 & 0.655\\
\hline

10 & \begin{tabular}[c]{@{}l@{}}   
        If you lack the will for such \\self-discipline, engage yourself in my work,\\ for selfless service can lead you at\\ last to complete fulfillment.  
        
\end{tabular} & \begin{tabular}[c]{@{}l@{}}   If you are also unequal to this method \\of constant practice, concentrate on  service\\ for Me; even thus serving Me you \\shall attain perfection.     

\end{tabular} & \begin{tabular}[c]{@{}l@{}}  And if you are not strong enough\\ to practise concentration, then devote yourself\\ to My service, do all yours acts for \\My sake, and you shall still \\attain the goal.  \end{tabular}  & 0.637 & 0.746 & 0.664\\
\hline

11 & \begin{tabular}[c]{@{}l@{}}   
      If you are unable to do even this,\\ surrender yourself to   me, disciplining yourself\\ and renouncing the results of all your actions. 
      
\end{tabular} & \begin{tabular}[c]{@{}l@{}}   If you are unable even to do this,\\ then dedicating all to Me, with mind  \\controlled, abandon the fruit of action.   

\end{tabular} & \begin{tabular}[c]{@{}l@{}} And if you are too weak even for \\this, then seek refuge in union with Me, \\and with perfect self-control renounce the fruit \\of your action. 

\end{tabular}  & 0.749 & 0.646 & 0.682\\
\hline

12 & \begin{tabular}[c]{@{}l@{}}   
      Better indeed is knowledge than mechanical \\practice. Better than knowledge is meditation. \\But better still is surrender of attachment to results,\\ because there follows immediate peace. 
      
\end{tabular} & \begin{tabular}[c]{@{}l@{}}   Better is knowledge than practice,\\ better than knowledge is concentration,\\  better than concentration is renunciation\\ of the fruit of all action, \\from which  directly issues peace.   

\end{tabular} & \begin{tabular}[c]{@{}l@{}}  Knowledge is superior to blind action,\\ meditation to mere knowledge,\\ renunciation of the fruit of action\\ to meditation, and where there is renunciation\\ peace will follow. 
\end{tabular}  & 0.673 & 0.854 & 0.681\\
\hline

13 & \begin{tabular}[c]{@{}l@{}}   
       That one I love who is incapable of ill\\ will, who is friendly and compassionate.\\ Living beyond the reach of I and mine\\ and of pleasure and pain, 
      
\end{tabular} & \begin{tabular}[c]{@{}l@{}}    Who has ill-will towards none, who is\\ friendly and compassionate, who has  shed all\\ thought of ‘mine' or ‘I', who regards\\ pain and pleasure alike, who is  long-suffering;      

\end{tabular} & \begin{tabular}[c]{@{}l@{}}  He who is incapable of hatred towards any being,\\ who is kind and compassionate, free from \\selfishness, without pride, equable \\in pleasure and in pain, and forgiving, 
\end{tabular}  & 0.558 & 0.811 & 0.507\\
\hline

14 & \begin{tabular}[c]{@{}l@{}}   
       patient, contented, self-controlled, firm \\in faith, with all their heart and \\all their mind given to me  with \\such as these I am in love. 
      
\end{tabular} & \begin{tabular}[c]{@{}l@{}}     Who is ever content, gifted with yoga,\\ self-restrained, of firm conviction,  who has \\dedicated his mind and reason to Me—that\\ devotee (bhakta) of Mine  is dear to Me.   

\end{tabular} & \begin{tabular}[c]{@{}l@{}}  Always contented, self-centred, self-controlled,\\ resolute, with mind and reason dedicated to Me,\\ such a devotee of Mine is My beloved. 

\end{tabular}  & 0.495 & 0.646 & 0.768\\
\hline

15 & \begin{tabular}[c]{@{}l@{}}   
         Not agitating the world or by it agitated,\\ they stand above the sway of elation,\\ competition, and fear: that one is my beloved.  
      
\end{tabular} & \begin{tabular}[c]{@{}l@{}}      Who gives no trouble to the world,\\ to whom the world causes no trouble,\\  who is free from exultation, resentment, fear\\ and vexation,—that man is dear to  Me.   

\end{tabular} & \begin{tabular}[c]{@{}l@{}}  He who does not harm the world,\\ and whom the world cannot harm,\\ who is not carried away by any impulse of joy,\\ anger or fear, such a one is My beloved. 

\end{tabular}  & 0.567 & 0.758 & 0.621\\
\hline

16 & \begin{tabular}[c]{@{}l@{}}   
       They are detached, pure, efficient,\\ impartial, never anxious, selfless in all their\\ undertakings; they are my devotees, very dear\\ to me.   
      
\end{tabular} & \begin{tabular}[c]{@{}l@{}}     Who expects nothing, who is pure, resourceful,\\ unconcerned, untroubled,  who indulges in no undertakings,\\—that devotee of Mine is dear to Me.     

\end{tabular} & \begin{tabular}[c]{@{}l@{}}  He who expects nothing, who is pure,\\ watchful, indifferent, unruffled, and who renounces\\ all initiative, such a one is My beloved. 

\end{tabular}  & 0.684 & 0.847 & 0.569\\
\hline

17 & \begin{tabular}[c]{@{}l@{}}   
        That one is dear to me who runs not\\ after the pleasant or away from the painful,\\ grieves not, lusts not, but lets things come\\ and go as they happen.  
      
\end{tabular} & \begin{tabular}[c]{@{}l@{}}      Who rejoices not, neither frets nor grieves,\\ who covets not, who abandons  both good \\and ill—that devotee of Mine is dear \\to Me.        

\end{tabular} & \begin{tabular}[c]{@{}l@{}}   He who is beyond joy and hate,\\ who neither laments nor desires,\\ to whom good and evil fortunes are the same,\\ such a one is My beloved. 

\end{tabular}  & 0.628 & 0.678 & 0.597\\
\hline

18 & \begin{tabular}[c]{@{}l@{}}   
       That devotee who looks upon friend and foe\\ with equal regard, who is not buoyed up\\ by praise nor cast down by blame, alike in\\ heat and cold, pleasure and pain, free \\from selfish attachments, 
      
\end{tabular} & \begin{tabular}[c]{@{}l@{}}     Who is same to foe and friend,\\ who regards alike respect and disrespect,\\  cold and heat, pleasure and pain,\\ who is free from attachment;   

\end{tabular} & \begin{tabular}[c]{@{}l@{}}    He to whom friend and foe are alike,\\ who welcomes equally honour and dishonour,\\ heat and cold, pleasure and pain, \\who is enamoured of nothing, 

\end{tabular}  & 0.827 & 0.885 & 0.885\\
\hline

19 & \begin{tabular}[c]{@{}l@{}}   
      the same in honor and dishonor,\\ quiet, ever full, in harmony everywhere,\\ firm in faith  such a one is dear to me. 
      
\end{tabular} & \begin{tabular}[c]{@{}l@{}}    Who weighs in equal scale blame and praise,\\ who is silent, content with  whatever his lot,\\ who owns no home, who is of steady mind,—\\that devotee of  Mine is dear to Me.    

\end{tabular} & \begin{tabular}[c]{@{}l@{}}   Who is indifferent to praise and censure,\\ who enjoys silence, who is contented \\with every fate, who has no fixed abode,\\ who is steadfast in mind, and filled \\with devotion, such a one is My beloved. 

\end{tabular}  & 0.541 & 0.732 & 0.583\\
\hline

20 & \begin{tabular}[c]{@{}l@{}}   
     Those who meditate upon this immortal dharma \\as I have declared it, full of faith and seeking\\ me as lifes supreme goal, are truly my devotees,\\ and my love for them is very great. 
      
\end{tabular} & \begin{tabular}[c]{@{}l@{}}    They who follow this essence of dharma,\\ as I have told it, with faith,  keeping Me as \\their goal,—those devotees are exceeding \\dear to Me.

\end{tabular} & \begin{tabular}[c]{@{}l@{}}  Verily those who love the spiritual wisdom \\as I have taught, whose faith never fails,\\ and who concentrate their whole nature on Me,\\ they indeed are My most beloved.” 

\end{tabular}  & 0.840 & 0.684 & 0.664\\
\hline

 * & & & & 0.609 (0.120) & 0.724 (0.096) & 0.633 (0.097) \\
\hline

\end{tabular}
}
 \caption{Semantic similarity of verses from Chapter 12. The cosine similarity (score) is given with the following combinations. Note that the comparisons are defined by Eknath Easwaran vs Mahatma Gandhi (Score 1),   Mahatma Gandhi vs Shri Purohit Swami (Score 2), and   Shri Purohit Swami vs Eknath Easwaran (Score 3). We provide the mean and standard deviation (in brackets) of the scores at the bottom (*).
 }
    \label{tab:semantic_similarity_chapter_12}
\end{table*}


\begin{table*}[htbp!]
\resizebox{\textwidth}{!}{%
\begin{tabular}{|l|l|l|l|l|l|l|}
\toprule
Chapter & Eknath Easwaran & Mahatma Gandhi & Shree Purohit Swami\\ [1ex]
\hline
1 & \begin{tabular}[c]{@{}l@{}} arjuna, krishna, battle, dharma, brahmins\\dhritarashtra, war, devadutta, bhishma, bhima

\end{tabular} & \begin{tabular}[c]{@{}l@{}} arjuna, maharatha,
battle, devadatta, armies,\\ sanjaya, bhima, kripa, yudhishthira, war

\end{tabular} & \begin{tabular}[c]{@{}l@{}} arjuna, dharmaraja, maharaja, battle, dhritarashtra,\\ devadatta, war, duryodhana, sanjaya, kripa \end{tabular} \\ [2ex]
\hline 

2 & \begin{tabular}[c]{@{}l@{}} arjuna, sorrow, souls, immortality, krishna,\\
death, arise, reverence, despair, disciple

\end{tabular} & \begin{tabular}[c]{@{}l@{}} arjuna, anguish, krishna, sorrowing, reverence, \\arise, heartedness, compassion, endure, slain

\end{tabular} & \begin{tabular}[c]{@{}l@{}} arjuna, equanimity, anguish, krishna, reverence, \\compassion, endure, terror, disciple, battle \end{tabular} \\ [2ex]
\hline 

3 & \begin{tabular}[c]{@{}l@{}} arjuna, devotion, wisdom, selfless, selflessly, \\krishna, knowledge, inaction, desires, brahman

\end{tabular} & \begin{tabular}[c]{@{}l@{}} arjuna, krishna, sacrifice, inaction,
detachment, \\renunciation, desires, pervading, yoga, prakriti

\end{tabular} & \begin{tabular}[c]{@{}l@{}} arjuna, wisdom, sacrifice, meditates,
inaction, \\krishna, meditate, worship, refraining, honour \end{tabular} \\ [2ex]
\hline

4 & \begin{tabular}[c]{@{}l@{}} teachings, arjuna, dharma, krishna, yoga,\\ eternal,
devotee, sages, changeless, surrendering

\end{tabular} & \begin{tabular}[c]{@{}l@{}} arjuna, yoga, wrath, sages, devotee,\\ essence, changeless, succession, learnt, varnas

\end{tabular} & \begin{tabular}[c]{@{}l@{}} arjuna, reincarnate, wisdom, krishna,
worship,\\ materialism, devotee, changeless, ancestors, births \end{tabular} \\ [2ex]
\hline 

5 & \begin{tabular}[c]{@{}l@{}} selfless, wisdom, renunciation, arjuna, krishna,\\ judgment, desires, sannyasa, duality, selfishly

\end{tabular} & \begin{tabular}[c]{@{}l@{}} renunciation, yoga, soul, sankhyas, virtue, actions,\\ peace, karmayoga, attain, ascetic

\end{tabular} & \begin{tabular}[c]{@{}l@{}} meditating, wisdom, renunciation, krishna, arjuna,\\ praise, divine, sin, purification, perceives \end{tabular} \\ [2ex]
\hline

6 & \begin{tabular}[c]{@{}l@{}} meditation, yoga, selfrealization, arjuna,
selfless,\\ attains, energy, abiding, refrain, renounce

\end{tabular} & \begin{tabular}[c]{@{}l@{}} yoga, yogi, arjuna, oneself,
contentment, nirvana,\\ purification, spirit, motionless, sannyasin

\end{tabular} & \begin{tabular}[c]{@{}l@{}} spirituality, meditation,
krishna, arjuna, celibacy,\\ purification, renounces, virtuous, vow, sage \end{tabular} \\ [2ex]
\hline

7 & \begin{tabular}[c]{@{}l@{}} arjuna, devotion, krishna, spiritual, prakriti,\\ eternal, universe, yoga,nature,realize

\end{tabular} & \begin{tabular}[c]{@{}l@{}} arjuna,vedas,transcending,yoga,
jiva,\\essence,evil,universe,manliness,mystery

\end{tabular} & \begin{tabular}[c]{@{}l@{}} arjuna,krishna,godless,spiritual, righteousness,\\
eternal,scriptures,beings,natures,unto \end{tabular} \\ [2ex]
\hline

8 & \begin{tabular}[c]{@{}l@{}} brahman,arjuna,devotion, krishna,
eternal,\\ souls, adhiyajna, death, existence, realize

\end{tabular} & \begin{tabular}[c]{@{}l@{}} brahman, arjuna, vedas, adhyatma, devotion,\\death,embodied,abode,yogi,departs

\end{tabular} & \begin{tabular}[c]{@{}l@{}} arjuna,devotion,consciousness,
eternity,souls,\\divinity,krishna,meditate,scriptures,omniscient \end{tabular} \\ [2ex]
\hline

9 & \begin{tabular}[c]{@{}l@{}} devotion, krishna, arjuna, eternal,
souls,\\ divine, behold, vijnana, wisdom, existence

\end{tabular} & \begin{tabular}[c]{@{}l@{}} arjuna,dharma, transcendent,
prakriti, doctrine,\\ souls, homage, multitude, condemn, lord

\end{tabular} & \begin{tabular}[c]{@{}l@{}} devotion, arjuna, krishna, godless,mysticism,souls,\\multitude,immovable,beings,universe \end{tabular} \\ [2ex]
\hline

10 & \begin{tabular}[c]{@{}l@{}} krishna, arjuna,brahman,devotion,sages,\\
creation,divine,manifestations,wisdom,equanimity

\end{tabular} & \begin{tabular}[c]{@{}l@{}} arjuna,brahman,devotion,gods,
beings,\\discernment,yoga,everlasting,contentment, inward

\end{tabular} & \begin{tabular}[c]{@{}l@{}} krishna,arjuna,devotion,
godless,wisdom,\\eternal,mankind,holiest,progenitors,contentment \end{tabular} \\ [2ex]
\hline

11 & \begin{tabular}[c]{@{}l@{}} arjuna,krishna,creation,behold,infinite,\\
celestial,perceive,yoga,exalted,sages

\end{tabular} & \begin{tabular}[c]{@{}l@{}} arjuna,krishna,gods,behold,
universe,\\multitudes,vasus,yoga,vestments,imperishable

\end{tabular} & \begin{tabular}[c]{@{}l@{}} arjuna,krishna,omnipresent,
universe,behold,\\celestial,embraced,radiance,powers,petal \end{tabular} \\ [2ex]
\hline

13 & \begin{tabular}[c]{@{}l@{}} devotion,brahman,krishna,buddhi,spiritual,\\ immortality, wisdom, arjuna,
sages, arise

\end{tabular} & \begin{tabular}[c]{@{}l@{}} brahman, consciousness,
devotion, arjuna, uprightness,\\ body, aphoristic, aversion, immovable,
pretentiousness

\end{tabular} & \begin{tabular}[c]{@{}l@{}} devotion, wisdom, vitality,
perception, krishna,\\ omniscient, nature, arjuna, matter, renunciation \end{tabular} \\ [2ex]
\hline

14 & \begin{tabular}[c]{@{}l@{}} krishna, arjuna, tamas, indolence, transcends,\\ reborn, wisdom,death,rajas,wombs

\end{tabular} & \begin{tabular}[c]{@{}l@{}} arjuna,tamas,sages,creation,
prakriti,\\attains,heedlessness,sattva,beings,restlessness

\end{tabular} & \begin{tabular}[c]{@{}l@{}} purity,indolence,krishna,
divinity,sinless,\\infatuation,spirit,reborn,eternal,knowledge \end{tabular} \\ [2ex]
\hline

15 & \begin{tabular}[c]{@{}l@{}} krishna,eternal,scriptures,wisdom,
essence,\\immutable,ashvattha,sages,tree,universe

\end{tabular} & \begin{tabular}[c]{@{}l@{}} vedanta,ashvattha,eternal,
souls,perceive,\\tree,ramified,abode,inward,ancient

\end{tabular} & \begin{tabular}[c]{@{}l@{}} eternal,creation,krishna,
scriptures,wisdom,\\perceive,celestial,regeneration,nourishment,tree \end{tabular} \\ [2ex]
\hline

16 & \begin{tabular}[c]{@{}l@{}} krishna,arjuna,demonic,scriptures,
purity,\\god,renunciation,qualities,hypocrisy,shall

\end{tabular} & \begin{tabular}[c]{@{}l@{}} arjuna,sacrifice,purity,
souls,wrath,spiritedness,\\devilish,aversion,pretentiousness,god

\end{tabular} & \begin{tabular}[c]{@{}l@{}} godless,krishna,wrath,
scriptures,harmlessness,\\modesty,shall,destroy,delusion,hypocritical \end{tabular} \\ [2ex]
\hline

17 & \begin{tabular}[c]{@{}l@{}} tamasic,krishna,sacrifices,worship,arjuna,\\scriptures,disregard,rajas,
hypocrisy,sattva

\end{tabular} & \begin{tabular}[c]{@{}l@{}} worship,krishna, yakshas, arjuna,forsake,\\shastra,sattva,rajas,pretentiousness,unholy

\end{tabular} & \begin{tabular}[c]{@{}l@{}} sacrifice,purity,krishna,
arjuna,scriptures,\\nourishing,aryans,vigour,unclean,covet \end{tabular} \\ [2ex]
\hline

18 & \begin{tabular}[c]{@{}l@{}} arjuna,discipline,sacrifice,renunciation,
sannyasa,\\discomfort,divine,tyaga,selfish,meaning

\end{tabular} & \begin{tabular}[c]{@{}l@{}} arjuna,sacrifice,renunciation,
doctrine,undertakes,\\tyagi,abandon,embodied,actions,wise

\end{tabular} & \begin{tabular}[c]{@{}l@{}} arjuna,benevolence,sacrifice,
renunciation,philosophy,\\relinquishment,desire,renounce,forgoing,act \end{tabular} \\
\hline

\end{tabular}
}
 \caption{Keywords for All chapters}
    \label{tab:keywords_all_chapters}
\end{table*}


\begin{table*}[!htb]
\small
\centering 
\begin{tabular}{| c | c | c | c | c | c |} 
\hline\hline 
 Eknath Easwaran & Score & Mahatma Gandhi & Score & Shri Purohit Swami & Score \\ [1ex] 
\hline 
devotion & 1.83 & devotion & 2.17 & devotee & 1.54\\ [0.5ex] \hline 
arjuna & 1.61 & arjuna & 1.54 & worship & 1.51 \\ [0.5ex] \hline
krishna & 1.52 & omnipresent & 1.20 & krishna & 1.33 \\ [0.5ex] \hline
yoga & 1.31 & yoga & 1.16 & arjuna & 1.29 \\ [0.5ex] \hline
selfless & 1.29 & mortals & 1.00 & eternal & 1.10 \\ [0.5ex] \hline
eternal & 1.22 & embodied & 0.98 & meditate & 0.99
 \\ [0.5ex] \hline
serene & 1.10 & lord & 0.93 & renunciation & 0.90 \\ [0.5ex] \hline
renouncing & 0.89 & unchanging & 0.92 & verily & 0.76 \\ [0.5ex] \hline
beings & 0.86 & attain & 0.87 & realise & 0.70\\ [0.5ex] \hline
established & 0.66 & steadily & 0.73 & attention & 0.60 \\ [1ex] \hline

\end{tabular}
\caption{Keywords extracted from Chapter 12 for the respective translations.}
\label{table:keywords_chapter_12} 
\end{table*}

\section{Further Analysis and Discussion}

We note that the Bhagavad Gita captures the conversation between Arjuna and Lord Krishna which takes place in the middle of the Kurushetra battlefield just before the war. The conversation is merely a few hours long and over the course of the Bhagavad Gita, Arjuna goes from being extremely distressed to a more calm and lucid state aware of his responsibilities. We wish to explore if the verses in which Arjuna has spoken can capture the variations of the sentiments expressed. Therefore, we provide an analysis where we calculate a sentiment polarity score in the range [-1,1] based on grouping the optimistic and pessimistic sentiments. In order to quantify the polarity of predicted sentiments, we assign a sentiment polarity score based on set of verses in each chapter by the respective speakers (Lord Krishna and Arjuna). We assign the sentiments such as "optimistic", "thankful" and "empathetic"  a score of +1, and further assign "pessimistic", "anxious", "sad", "annoyed", and "denial"  a score of -1. Finally, we assign   "joking" and "surprise",   a score of 0 since they can have positive or negative connotation. It should be noted that Arjuna does not speak in certain chapters (7, 9, 13, 15, 16) and hence, the polarity score is zero.
 We show the chapter-wise polarity score for Eknath Easwaran's translation in Figure \ref{arjuna_sentiments} for   Arjuna  where we observe that Chapter 1 has the maximum negative sentiment over time (chapters). This is because in the beginning of the Bhagavad Gita,  Arjuna is distraught and refuses to fight the war on seeing his family and friends on the opposite side. Moreover, we notice the polarity increases towards positive  sentiments from Chapter 2 and then there is a bump in Chapter 11. This is because in Chapter 11, Arjuna is overwhelmed by Lord Krishna's \textit{universal form} (viraat roopa)  and considers himself fortunate to have witnessed it, hence the maximum positive sentiments  come from this chapter.  Similarly, we have shown the variation of Lord Krishna's sentiments in Figure \ref{krishna_sentiments}, where he begins with a neutral sentiment polarity  and then becomes negative in Chapter 2, and positive afterwards till Chapter 13. Moving on, his sentiments fluctuates  and also becomes close to neutral as the time progresses.

  In  Table \ref{tab:chapter6_verses}, verse 2 validates our earlier observation of the differences in central themes (selfless service - Eknath Easwaran, right action - Shri Purohit Swami). It is important to note that the BERT model has been trained on a dataset of tweets which featured sarcastic comments and jokes. Hence, the Senwave dataset contains the label - "joking". However, in the Bhagavad Gita, this sentiment is not apt and therefore, some verses may have been classified as "joking" inaccurately. Please note that we have shown the pre-processed verses here as they have been used for sentiment prediction and not the original verses.

   In the heatmaps (Figure \ref{fig:heatmaps}), we observe that verses labelled as 'joking' have a significant overlap with "optimistic", "annoyed" and "surprise". We could have removed the joking label from the training data as we know that the Bhagavad Gita does not contain jokes, but it is interesting to know that due to the translation style, humour was detected in the way certain questions were asked and answered. We present some  verses in Table \ref{tab:joking_verses} that were classified as "joking".  We note that these were not jokes but verses that can be considered humorous, which is based on the way experts have hand-labelled tweets during COVID-19 outbreak in 2020. We can note this as a major limitation since we are using hand labelled sentiments from social media as training data and applying it to a entirely different area - philosophy and religion.

We have two major limitations of the proposed framework. The first major limitation is in terms of  the model and the training data, since the model is based on pre-trained BERT model which is generic and cannot capture the essence of Hindu philosophy. We note that there are specific Sanskrit terms that are difficult to translate. There are a number of BERT specific models, for different domains such as Med-BERT for   medicine  \cite{rasmy2021med} and Legal-BERT for  law \cite{chalkidis2020legal}. In the case of philosophy, we do not have any  pre-trained model for masked language modelling. Our application is unique, and hence a BERT-based on   western philosophy will not suffice. We would need a pre-trained  BERT model  for Hindu philosophy in order to capture the essence and specific words such as \textit{Brahman, avatar, karma, bhagwan, Vishnu, maya, viraat roopa, and dharma}. Once the key terms are recognised,    pre-processing of the respective texts will not be crude where we simply change the words to English without considering the context. We need to note that the English language is heavily influenced by the concept of God in Abrahamic religions (Christianity, Judian and Islam); hence, the term God has a specific meaning which is in contrast to equivalent concepts such as  Brahman or Vishnu. The Abrahmic notion of God refers to the creator who is not part of the universe \cite{van2008nature}, but can control the universe and controls the mortals (humans) with rewards and punishment. The Hindu definition of God is entirely different \cite{vivekananda1937essentials}, where the creator and the creation are seen as the same. Hindu philosophy views the creation featuring two major aspects; i.e consciousness (Purusha)  which is depicted by the male Gods such as Brahma and nature and matter; nature (Prakrati) which is depicted by their consorts,  the Goddesses such as Saraswati, Parvati and Laxmi.

 We note that in Chapter 2 of  the Bhagavad Gita, there are fifteen predicates  used to define the Atman which in some philosophical traditions such as the Advaita Vedanta is seen same as the  Paramatman (supreme Atman) and also  Brahman (ultimate reality). The descriptions of Atman are given by the terms avinasin (non-perishable), avyaya (non-changeable), anasin (non-destructible), aprameya (nonmeasurable), aja, (non-born) nitya (eternal), sasvata (constant), sarvagata (omnipresent), sthanu (steadfast) acale (non-movable),
sanatana (everlasting) avyakta (non-manifest),  acintya (nonconceivable), avikarya (non-variable),  and avadhya (non-violable).
These definitions of the supreme reality are also found in the Upanishads \cite{upanishads_swami_paramananda}. These include, 
tat tvam asi (thou art that) - Chandogya Upanishad - 6.8.7
aham brahmāsmi (I am Brahman) - Brihadaranyaka Upanishad 1.4.10  which build the foundation of Advaita Vedanta \cite{parimala2017unit}.
 
  Hinduism is often labelled as polytheist, but in principle it is a form of pantheism \cite{o1935popular} which is a philosophical system based on the notion  that reality is identical with divinity, i.e the creator and creation are the same. Apart from this, Hindu philosophy also caters for schools of atheism, such as Carvaka philosophy \cite{chatterjee2016introduction}; and non-theism, such as Buddhism and Jainism. These schools and definitions make it difficult to translate Sanskrit words to English since they have historical development of the terms with philosophical annotations. For examples, Vishnu means "preserver", "vastness" and Brahma "refers to   "the creator" and Shiva can refer to "the auspicious one"  and depending on the content, various different names are used. For instance,  Natraja refers to the the dancing Shiva -  widely known as the cosmological dancer \cite{beltz2011dancing}. Rudra often refers to the personality of Shiva in anger.  Moreover, certain Sanskrit terms with similar wordings can have entirely different meaning, e. Bhram (illusion), Brahma (creator), Brahman (ultimate reality), Brahmand (universe), and Brahmin (priest or academic caste/profession).
  
Apart from the context and the difference in the philosophical system, we note that the challenge of translating the Bhagavad Gita is not only that it is written in an ancient language which is not a conversational language nowadays, but also that it is written as a poem (as a song). The term 'gita' translates to 'song' and Bhagavad Gita translates to 'song of the divine'. The Bhagavad Gita has been written as a song so that it can be remembered and sung and this is how it has been transmitting through oral tradition. It is well known that translating poetry and songs leads to losing rhyming patterns and at times meaning, since  certain metaphors and expressions are constrained to specific languages, cultural traditions and time-frames \cite{low2003translating,jones2011poetry}. Moreover, certain verses in the Gita can have multiple layers of meaning since they are written in the form of poetry and feature literary devices such as metaphors \cite{dupriez1991dictionary}. Hence, translating such texts is very challenging, due to language and style of the ancient Hindu writers also known as Rishi's  \cite{elizarenkova1995language,krishnamoorthy1991poetic}. Note there are certain prominent translations, such as the 'Bhagavadgita' by Sir Edwin Arnold that give a poetic interpretation \cite{arnold1885song} but do not distinguish the different verses. A recent translation by  Badhe \cite{sushrat2015x}
  maintained the verses with rhythm and rhyming patterns  so that it  does not lose the essence of the song-verse style from  the original Sanskrit version. Future work can be done by analysis of sentiment and semantic similarity of translations that maintain rhythm and rhyme.

 \begin{table*}[htbp!]
\resizebox{\textwidth}{!}{%
\begin{tabular}{|l|l|l|l|l|l|l|l|}
\toprule
Chapter &  & Eknath Easwaran & Mahatma Gandhi & Shri Purohit Swami\\ [1ex]
\hline

5 & Verse 23 & \begin{tabular}[c]{@{}l@{}} But those who overcome the\\ impulses of lust and anger which\\ arise in the body are made \\whole and live in joy.	

\end{tabular} & \begin{tabular}[c]{@{}l@{}} The man who is able even here on\\ earth, before he is released from\\ the body, to hold out against the floodtide\\ of lust and wrath,—he is a yogi,\\ he is happy.

\end{tabular} & \begin{tabular}[c]{@{}l@{}} He who, before he leaves \\ his body, learns to surmount \\ the promptings of desire and anger \\ is a saint and is happy.	 \end{tabular} \\
\hline 

 & Predicted Sentiment & \begin{tabular}[c]{@{}l@{}} Joking, Optimistic

\end{tabular} & \begin{tabular}[c]{@{}l@{}} Joking, Optimistic

\end{tabular} & \begin{tabular}[c]{@{}l@{}} Joking, Optimistic \end{tabular} \\
\hline \hline

11 & Verse 14 & \begin{tabular}[c]{@{}l@{}} Filled with amazement, his hair \\standing on end in ecstasy, he bowed\\ before the Lord with joined palms and \\spoke these words.

\end{tabular} & \begin{tabular}[c]{@{}l@{}} Then Arjuna, wonderstruck and thrilled\\ in every fibre of his being, bowed\\ low his head before the Lord, addressing\\ Him thus with folded hands. Arjuna Said:	

\end{tabular} & \begin{tabular}[c]{@{}l@{}} Thereupon Arjuna, dumb with awe,\\ his hair on end, his head bowed,\\ his hands clasped in salutation, addressed the\\ Lord thus: \end{tabular} \\
\hline 

 & Predicted Sentiment & \begin{tabular}[c]{@{}l@{}} Joking

\end{tabular} & \begin{tabular}[c]{@{}l@{}} Joking, Surprise

\end{tabular} & \begin{tabular}[c]{@{}l@{}} Joking, Surprise \end{tabular} \\
\hline \hline

11 & Verse 27 & \begin{tabular}[c]{@{}l@{}} All are rushing into your\\ awful jaws; I see some of them\\ crushed by your teeth.

\end{tabular} & \begin{tabular}[c]{@{}l@{}} Are hastening into the fearful\\ jaws of Your terrible mouths. Some\\ indeed, caught between Your teeth,\\ are seen, their heads being\\ crushed to atoms.

\end{tabular} & \begin{tabular}[c]{@{}l@{}} I see them all rushing headlong\\ into Your mouths, with terrible tusks,\\ horrible to behold. Some are mangled between\\ your jaws, with their heads crushed\\ to atoms. \end{tabular} \\

\hline

& Predicted Sentiments & \begin{tabular}[c]{@{}l@{}} Joking

\end{tabular} & \begin{tabular}[c]{@{}l@{}} Joking, Anxious, Annoyed

\end{tabular} & \begin{tabular}[c]{@{}l@{}} Joking \end{tabular} \\

\hline
\hline

\end{tabular}
}
 \caption{Selected verses with 'joking' as the predicted sentiment.}
    \label{tab:joking_verses}
\end{table*}

\section{Conclusion and Future work} 
 
We presented a framework for comparing translations  of the Bhagavad Gita using sentiment and semantic analysis. 
In terms of the comparison of the translations with statistical measures (such as bigrams and trigrams), we found major differences in the translations. Due to the nature of the Sanskrit language and the fact that the Bhagavad Gita is a song with rhythm and rhyme, it is not surprising that different translators used different vocabulary to describe the same concepts. Moreover, we note the vast difference in the dates of the translations (1935, 1946, and 1985); hence it is natural to find a major change in vocabulary and expressions. Moreover, we also found that two of the earlier  translators (Mahatma Gandhi and Shri Purohit Swami) used archaic words which were pre-processed to modern vocabulary so that it is easier for modelling. In the translation by Eknath Easwaran, we did not find any archaic words. Furthermore, the translators also used different names for Arjuna and Lord Krishna, which we also pre-processed. Hence, the language and vocabulary across the the different translations vastly differ. 

In the sentiment analysis, we find that although the language and vocabulary has been vastly different, the BERT-based language model has been able to recognise the sentiments well which have been mostly similar given the different chapters of the respective translations. The sentiments such as optimistic, annoyed and surprised have been the most expressed while the model also considered a significant number of verses as joking.  It is also interesting to find how the sentiment polarity changes for Arjuna and Lord Krishna over time, as the conversation deepens where Arjuna is pessimistic towards the beginning and then becomes optimistic with knowledge of Hindu philosophy and Karma Yoga imparted by Lord Krishna. Apart from the knowledge given by the philosophy of karma and dharma, the polarity of the sentiments expressed which changed over time  could be a  reason why the Bhagavad Gita is seen as a book of psychology, management and conflict resolution. The sentiments expressed by Krishna shows that with philosophical knowledge of dharma and mentor-ship, a troubled mind can get clarity for making the right decisions in times of conflict. 

The results from the semantic analysis also showed that the meaning conveyed across the different translations were similar although different language and vocabulary were used. Further, the keywords extracted from the different translations also showed similar terms, although in different order of importance.

Future work can focus in using Sanskrit expert translators to review the sentiments presented by the framework for the different translators. Moreover, we can also apply the framework to translations of the Bhagavad Gita that retain rhythm and rhyme. The same framework can be used to analyse the translations of other texts, these include the Upanishads and   also translations of texts in other domains from different languages.

\section*{Data and Code}

Python-based open source code and data can be found here
\footnote{\url{https://github.com/sydney-machine-learning/sentimentanalysis_bhagavadgita}}.
 
\section*{Author contributions statement}

R. Chandra devised the project with the main conceptual ideas and experiments and contributed to overall writing,  literature review and discussion of results.  V. Kulkarni  provided implementation and experimentation and further contributed in results visualisation and analysis. 

\section*{Acknowledgement}

We acknowledge contributions of Sweta Rathi who supported earlier part of the project.

\bibliographystyle{IEEEtran}

\bibliography{language,usyd,sample,sample_,2020June,covid,Hinduism}


\EOD

\end{document}